\documentclass[10pt,twocolumn,letterpaper]{article}

\usepackage{iccv}            

\newcommand{\xmark}{\ding{55}}
\newcommand{\myfirstpara}[1]{{\par \noindent \textbf{#1.}}}
\newcommand{\mypara}[1]{\vspace{0em} \myfirstpara{#1}}

\def\sfda{\texttt{SFDA}\xspace}
\def\sfdaod{\texttt{SF-DAOD}\xspace}

\def\focalnet{\texttt{FocalNet-DINO}\xspace}

\def\vit{\texttt{VIT}\xspace}

\def\ema{\texttt{EMA}\xspace}
\def\vit{\texttt{VIT}\xspace}
\def\detr{\texttt{DETR}\xspace}
\def\fnd{\texttt{FND}\xspace}
\def\coco{\texttt{MS-COCO}\xspace}
\def\pascal{\texttt{PASCALVOC}\xspace}
\def\wsod{\texttt{WSOD}\xspace}
\def\uda{\texttt{UDA}\xspace}
\def\daod{\texttt{DAOD}\xspace}

\def\sfod{\texttt{SFOD}\xspace}
\def\soap{\texttt{SOAP}\xspace}
\def\lods{\texttt{LODS}\xspace}
\def\asfod{\texttt{A2SFOD}\xspace}
\def\irg{\texttt{IRG}\xspace}
\def\pets{\texttt{PETS}\xspace}
\def\sota{\texttt{SOTA}\xspace}

\def\cf{\texttt{C2F}\xspace}
\def\cb{\texttt{C2B}\xspace}
\def\kc{\texttt{K2C}\xspace}
\def\sc{\texttt{S2C}\xspace}
\def\ii{\texttt{In2IB}\xspace}

\def\ab{\texttt{A2B}\xspace}
\def\mi{\texttt{MI}\xspace}
\def\bcd{\texttt{BCD}\xspace}

\def\fpi{\texttt{FPI}\xspace}
\def\froc{\texttt{FROC}\xspace}
\def\ap{\texttt{AP}\xspace}
\def\map{\texttt{mAP}\xspace}
\def\vit{\texttt{VIT}\xspace}

\newcommand{\insightbox}[1]{%
    \begin{tcolorbox}[colframe=black!70, colback=blue!5, boxrule=1pt, arc=4mm]
        \includegraphics[width=0.4cm]{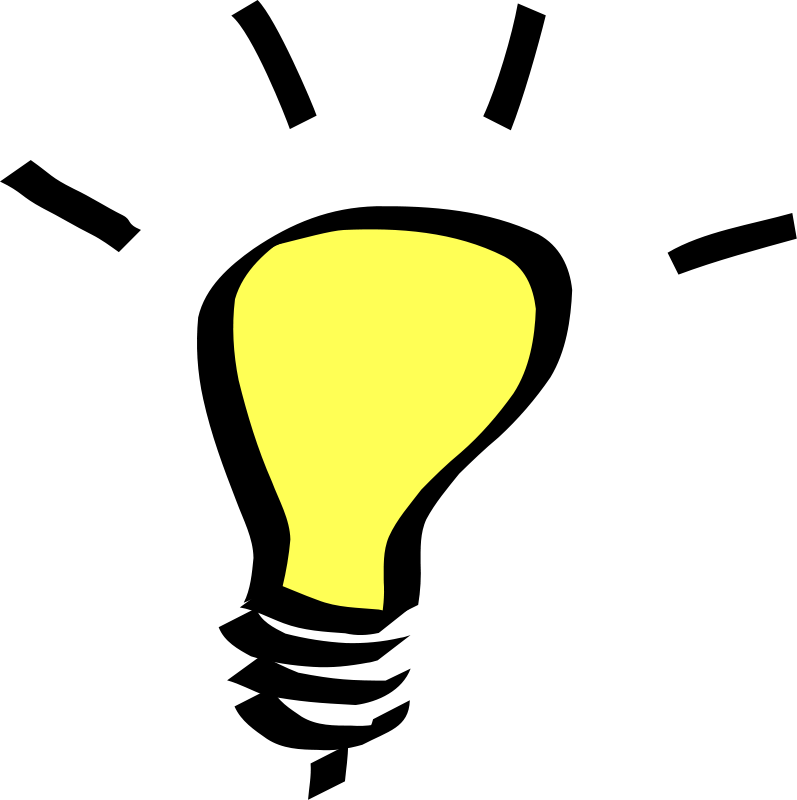}
        \textbf{\small#1}
    \end{tcolorbox}
}

\definecolor{iccvblue}{rgb}{0.21,0.49,0.74}
\usepackage[pagebackref,breaklinks,colorlinks,allcolors=iccvblue]{hyperref}
\usepackage[utf8]{inputenc} 
\usepackage[T1]{fontenc}    
\usepackage{hyperref} 
\usepackage{url}      
\usepackage{siunitx}
\usepackage{helvet}
\usepackage{booktabs}       
\usepackage{amsfonts}       
\usepackage{nicefrac}       
\usepackage{microtype}      
\usepackage{amsmath, amsthm, mathtools}
\usepackage{amsthm}  
\usepackage{amsmath} 
\usepackage{amssymb} %
\usepackage{changepage,threeparttable}
\usepackage{array,multirow,graphicx}
\usepackage{soul}
\usepackage[normalem]{ulem}
\useunder{\uline}{\ul}{}
\usepackage{subcaption}
\usepackage{booktabs}
\usepackage{xcolor}
\usepackage[font=small]{caption}
\usepackage{fontawesome}
\usepackage{pifont}
\usepackage{alphalph}
\usepackage[noend]{algpseudocode}
\usepackage{amssymb} 
\usepackage{tabularx}
\usepackage{xspace}
\usepackage{bm} 
\usepackage{array}
\usepackage{tcolorbox}  
\usepackage{graphicx}   
\usepackage{wrapfig}
\usepackage{subcaption}
\usepackage{booktabs}
\usepackage{adjustbox,makecell}
\usepackage[inline]{enumitem}
\usepackage{color, colortbl}
\usepackage{amsmath,algorithm}

\def\paperID{1966} 
\def\confName{ICCV}
\def\confYear{2025}

\title{TITAN: Query-Token based Domain Adaptive Adversarial Learning}

\author{Tajamul Ashraf\\
MBZUAI\\
Masdar City, Abu Dhabi, UAE\\
{\tt\small tajamul.ashraf@mbzuai.ac.ae}
\and
Janibul Bashir\\
National Institute of Technology Srinagar\\
Hazratbal, Jammu and Kashmir\\
{\tt\small janibbashir@nitsri.net}
}
\makeatletter
\let\@oldmaketitle\@maketitle
\renewcommand{\@maketitle}{
  \@oldmaketitle
  \begin{center}
    \vspace{-1em}
    {\large\textbf{This paper has been accepted at ICCV 2025.}}
  \end{center}
  \vspace{1em}
}
\makeatother

\begin{document}
\maketitle
\begin{abstract}
We focus on source-free domain adaptive object detection (\sfdaod) problem when source data is unavailable during adaptation and the model must adapt to unlabeled target domain. Majority of approaches for the problem employ a self-supervised approach using a student-teacher (\texttt{ST}) framework where pseudo-labels are generated via a source-pretrained model for further fine-tuning. We observe that the performance of a student model often degrades drastically, due to collapse of teacher model primarily caused by high noise in pseudo-labels, resulting from domain bias, discrepancies, and a significant domain shift across domains. To obtain reliable pseudo-labels, we propose a Target-based Iterative Query-Token Adversarial Network (\textbf{TITAN}) which separates the target images into two subsets that are similar to the source (easy) and those that are dissimilar (hard). We propose a strategy to estimate variance to partition the target domain. This approach leverages the insight that higher detection variances correspond to higher recall and greater similarity to the source domain. Also, we incorporate query-token based adversarial modules into a student-teacher baseline framework to reduce the domain gaps between two feature representations. Experiments conducted on four natural imaging datasets and two challenging medical datasets have substantiated the superior performance of \textbf{TITAN} compared to existing state-of-the-art (\sota) methodologies. We report an \map improvement of +22.7, +22.2, +21.1 and +3.7 percent over the current \sota on  \cf, \cb , \sc, and \kc benchmarks respectively.

\end{abstract}
\section{Introduction}
\label{introduction}
\myfirstpara{Object Detection}
Object detection is a well studied problem in computer vision \cite{carion2020end, dosovitskiy2020image, zhu2020deformable, zhang2022dino, yang2022focal}. The success of deep learning techniques for the problem has been supported by the abundance of extensively annotated detection datasets \cite{cordts2016cityscapes, everingham2010pascal, geiger2013vision, lin2014microsoft, yu2020bdd100k}, which facilitates the supervised training of deep object detection models.

\begin{figure}[t]
    \centering
    \includegraphics[width=\linewidth]{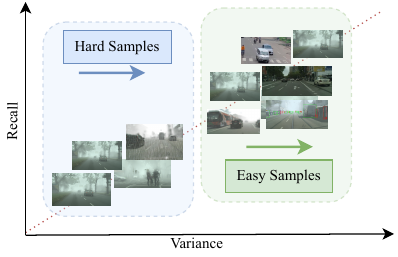}
      \caption{The core idea of our framework (\textbf{TITAN}) is that higher detection variances signal high recall and similarity to the source domain, enabling us to divide the target domain into easy and hard subsets.}
    \label{f1_teasor}
\end{figure}

\mypara{Unsupervised Domain Adaptation (UDA)}
It has been widely observed that, despite their effectiveness in familiar visual contexts, deep object detection models often struggle to generalize to new visual domains. Unsupervised Domain Adaptation (\uda) is a popular solution strategy \cite{chen2017no, ganin2016domain, hoffman2018cycada, hoffman2016fcns, lo2022learning, saito2018maximum, tzeng2017adversarial}, which bridges the gap between source and target domains by aligning the feature distributions \cite{chen2018domain, he2019multi, inoue2018cross, saito2019strong, vs2023instance, huang2022aqt, zhao2023masked}. However, the strategy requires access to source-domain data at the adaptation stage, which severely limits its applicability \cite{liang2020we, liu2021source, xia2021adaptive}. This motivates us to focus on Source-Free Domain Adaptive Object Detection (\sfdaod) in this work.

\mypara{Source-Free Domain Adaptive Object Detection (SF-DOAD)}
\sfdaod has received significant attention for the image classification task in recent years \cite{yang2022attracting, zhang2022divide, wang2022metateacher, jing2022variational, dong2021confident, huang2021model, yang2021exploiting}. However, there are relatively fewer works specifically addressing \sfdaod \cite{huang2021model,vs2023towards, oza2023unsupervised, liu2023periodically, chu2023adversarial}. Given the complexities of cluttered background, viewpoint variations, and many negative samples in an object detection problem, directly applying traditional \sfda methods for classification tasks to \sfdaod often leads to unsatisfactory performance. Thus, there is a need to develop techniques specially tailored for \sfdaod.

\mypara{Problems in Current SF-DOAD Approaches}
Many popular \sfdaod techniques utilize a self-supervised approach in a student-teacher (\texttt{ST}) framework. These approaches rely on self-training using pseudo-labels produced by a model pre-trained on the source domain. ~\cite{li2022source, xiong2021source,li2021free, chu2023adversarial, vs2022mixture}. However, if the source data is biased, or the domain shift between source and target domains is high, then there is significant noise in the pseudo-labels, which impacts the training of a student model \cite{deng2021unbiased}. Since the pseudo-label error is significant, hence, Exponential Moving Average (\ema) step which updates the teacher model from the student model's weight, ends up corrupting the teacher model as well. This is typically not a problem in the \uda setting, as supervised data from the source domain, acts as an anchor and prevents error accumulation in the student, and then the \ema step ensures that the teacher model does not get corrupted at any point in the training/adaptation.

\mypara{Solution Strategies to Mitigate SF-DOAD Problems}
To tackle the above issues in \sfdaod recent techniques \cite{oza2023unsupervised, chu2023adversarial} have proposed to use a larger update step size for \ema to slow down the teacher model's updating process deliberately. An alternative strategy involves emphasizing the past teacher model's influence by adjusting its contribution, thereby preserving previous knowledge and reducing the rate of model updates. However, such attempts have demonstrated limited effectiveness~\cite{liu2023periodically}.


\mypara{Our Insights and Proposed Strategy}
To tackle this issue, we propose a query-token-driven adversarial learning approach (\textbf{TITAN}). Our method employs a variance-based detection strategy to separate target data into easy and challenging subsets, leveraging the insight that greater detection variance aligns with higher recall and stronger resemblance to the source domain, as illustrated in Fig.~\ref{f1_teasor}.  Next, we integrate query-token-driven adversarial modules within a transformer-based student-teacher framework to bridge domain gaps in both local and instance-level feature representations, utilizing the \focalnet encoder and decoder accordingly.

\mypara{Contributions}
\begin{enumerate*}[label=\textbf{(\arabic*)}]
	\item We highlight the problem of training instability of a student model in \sfdaod setting due to incorrect pseudo labels from a source pretrained model in the presence of source data bias, or large domain shift between source and target domains.
	\item We identify a strong link between detection variance and resemblance to the source data. Leveraging this insight, we establish a method to categorize the target domain into easy and challenging subsets.
	\item We introduce a query-token based adversarial alignment approach to refine the feature space, ensuring the generation of reliable pseudo-labels for the student-teacher framework.
	\item We conduct extensive evaluation on four natural image adaptation benchmarks. We report an \map of 50.2, 38.3, 59.8, and 53.2 on \cf, \cb, \sc, and \kc benchmarks respectively, against the performance of 40.9, 31.6, 49.4, and 51.3 by the current \sota \cite{liu2023periodically, chu2023adversarial}.
    \item We observe that \sfdaod formulation is especially used for medical imaging scenarios where sharing source data is harder due to stricter privacy constraints, and where source data bias is more likely due to smaller datasets captured in a single center mode. Hence, we experiment and demonstrate state-of-the-art (\sota) results on cross-domain datasets for Breast Cancer Detection (\bcd) from mammograms. We report a recall of 0.78 and 0.51 at an \fpi of 0.3 on RSNA-BSD1K~\cite{carrrsna} to INBreast~\cite{moreira2012inbreast} and DDSM~\cite{lee2017curated} to INBreast~\cite{moreira2012inbreast} datasets respectively against the performance of 0.25 and 0.15 by the current \sota \cite{vs2023instance}.
     
\end{enumerate*}

\section{Related Work}
\label{related}

\textbf{Object Detection.}
In recent years, object detection, being a crucial computer vision task, has garnered considerable attention \cite{cheng2023towards, zou2023object}. Many object detection methods perform box regression and category classification using techniques such as anchors \cite{redmon2016you, lin2017focal,redmon2017yolo9000}, proposals \cite{ren2015faster, girshick2015fast}, and points \cite{wang2017point,zhou2019objects, tian2022fully}. Traditional object detection systems heavily rely on extensive datasets like \coco \cite{lin2014microsoft} and \pascal \cite{everingham2010pascal}, which necessitate significant time investment for annotation due to the large number of samples for each object category. Weakly supervised object detection (\wsod) methods \cite{zhang2020weakly,zhang2021weakly} leverage image-level labeled data \cite{ma2024codet} to train object detectors. In general, \wsod considers an image as a collection of region proposals and applies multiple instance learning to assign the image-level label to these proposals. \cite{bilen2015weakly,cinbis2016weakly, bilen2016weakly,wang2023alwod}. By using less expensive classification data, \wsod can expand the detection vocabulary without requiring costly instance-level annotations in object detection \cite{deng2009imagenet, zhou2022detecting, wang2023alwod,kniesel2023weakly, gungor2024boosting, gungor2024boosting, wang2024weakly}. More recently, transformer-based models \cite{ashish2017attention} have been developed for object detection, exploring token-wise dependencies for context modeling. Following the pioneering work of Vision Transformer (\vit) \cite{dosovitskiy2020image} and Detection Transformer (\detr) \cite{carion2020end}, transformers have emerged as a promising architecture in computer vision, demonstrating their efficacy in various tasks including object detection \cite{carion2020end, zhu2020deformable, zheng2020end, dai2021up, zhang2022dino, yang2022focal}. Whereas most approaches emphasize supervised learning settings, our goal is to improve the model's ability to adapt to unseen domains without requiring extra annotations. To achieve this, we employ FocalNet-DINO (\fnd) \cite{yang2022focal} as our baseline detection model, chosen for its streamlined architecture and state-of-the-art transfer learning capability.

\textbf{Source-Free Domain Adaptive Object Detection (SF-DOAD).}
In real-world applications, accessing source data during subsequent adaptation phases is often limited due to privacy laws, data transfer restrictions, or proprietary concerns, especially in medical imaging. Source-Free Domain Adaptive Object Detection (\sfdaod) enables knowledge transfer from pre-trained source models to the target domain without relying on sensitive source data \cite{luo2024crots}. Due to these constraints, \sfdaod has become more challenging compared to traditional Domain Adaptation Object Detection (\daod) methods, leading to the emergence of \sfdaod as a distinct branch within the domain adaptation framework in recent years. Given the intricate nature of object detection tasks, which involve numerous regions, multi-scale features, and complex network architectures, coupled with the absence of source data and target pseudo-labels, it is evident that straightforward application of existing Unsupervised Domain Adaptation (\uda) and Domain Adaptation Object Detection (\daod) methods is inadequate. The Source-Free Object Detection (\sfod) method proposed in \cite{li2021free} leverages self-entropy descent to generate reliable pseudo-labels for self-training. \soap \cite{xiong2021source} applies domain perturbations to the target data, enabling the model to learn domain-invariant features that remain robust against variations. \lods \cite{li2022source} incorporates a style enhancement module alongside a graph alignment constraint to promote the extraction of domain-agnostic features. \asfod~\cite{chu2023adversarial} classifies target images into source-similar and source-dissimilar groups, followed by adversarial alignment using \st models. \irg~\cite{vs2023instance} introduces an instance relation graph network combined with contrastive loss to enhance contrastive representation learning. \pets \cite{liu2023periodically} integrates a dynamic teacher model within the ST framework and introduces a consensus mechanism that merges predictions from both static and dynamic teacher models. Despite leveraging the ST framework, most of these approaches often overlook the challenge of overcoming local optima and training instability resulting from a single updated teacher model. 

\section{Methodology}
\label{methodology}
\begin{figure*}[h!]
    \centering
    \includegraphics[width=\textwidth]{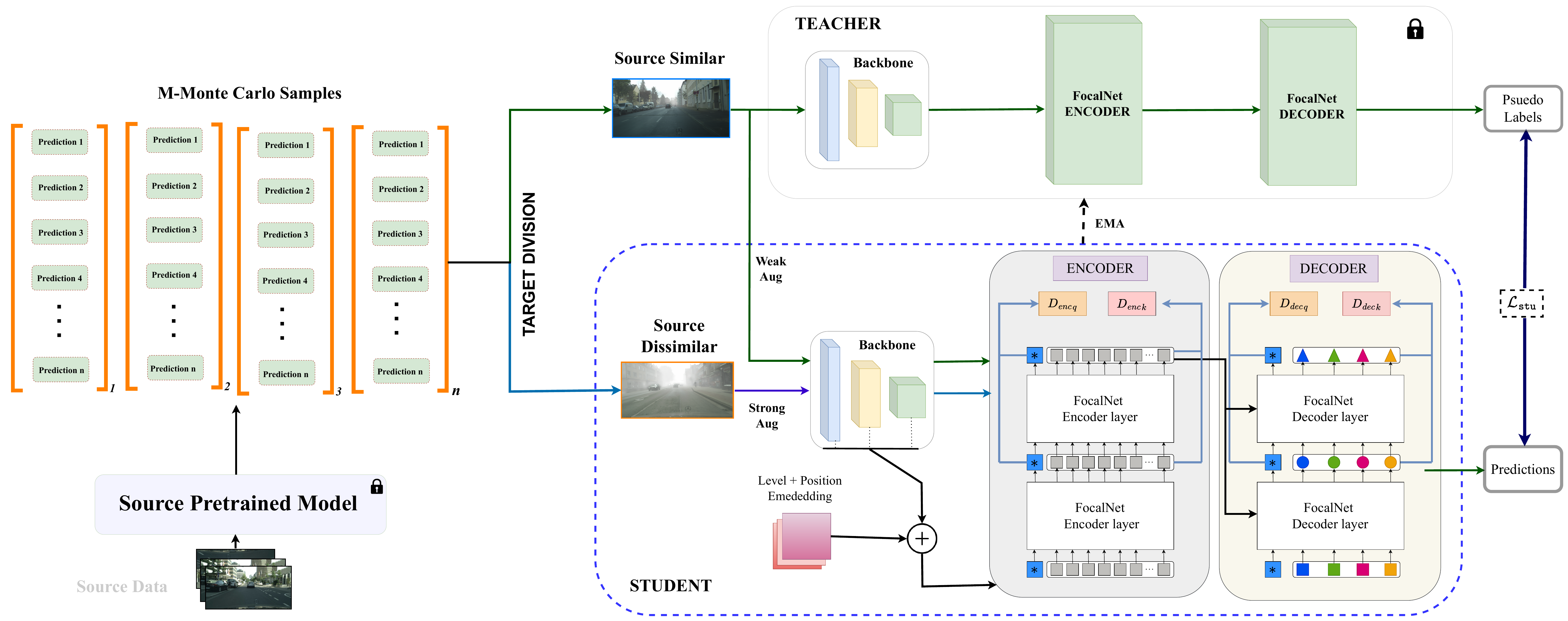}
    \caption{ Overview: Target-based Iterative Query-Token Adversarial Network (\textbf{TITAN})}
    \label{f2_architecture}
\end{figure*}

Source-free domain adaptive object detection (\sfdaod) aims to adapt a detector, initially trained on a source domain, to an unlabeled target domain without direct access to source data. Given an unlabeled target dataset $\{X_{i}^t\}_{i=1}^{N}$ (where $N$ denotes the total number of images) and a detector $F$ initialized with source-trained parameters $\theta_s$ (e.g., a \focalnet model~\cite{yang2022focal}), the objective is to refine these parameters to $\theta_t$ for effective performance in the target domain. In this work, we introduce a novel approach, \textbf{TITAN}, whose overall framework is depicted in Fig.~\ref{f2_architecture}. Our method categorizes target samples into two groups: source-similar (easy) and source-dissimilar (hard), using prediction variance from the source-trained detector \( F_{\theta_s} \). These groups are subsequently aligned using query-token-based adversarial learning within a \texttt{ST} framework. The following sections provide a detailed explanation of each stage.

\subsection{Target Domain Division}
Aligning source and target domains is common in traditional domain adaptation tasks~\cite{DBLP:conf/cvpr/Kang0YH19,DBLP:conf/cvpr/ZhuPYSL19,DBLP:conf/cvpr/0007CGV19}. While alignment can be performed in either data space~\cite{DBLP:conf/cvpr/ChenZD0D20} or feature space~\cite{DBLP:conf/cvpr/SaitoUHS19}, the absence of source data presents a unique challenge for domain alignment.

Even though the source data is inaccessible, the source-pretrained model retains crucial knowledge about the source domain. We introduce a strategy to partition the target dataset into two groups, leveraging the pre-trained model to establish an explicit source-target domain distinction. To achieve this, we define a detection variance-based criterion, where variances are computed from the pre-trained model’s predictions on the target samples. A higher variance suggests a stronger resemblance to the source domain. In particular, the model exhibits greater uncertainty (hard samples) when predicting source-similar images, leading to elevated variance values compared to source-dissimilar images. The detection variance is determined using the following formulation:

\begin{equation}
    v_i = \mathbb{E}[(F_{\theta_s}(X_i) - \mathbb{E}[F_{\theta_s}(X_i)])^2],
\end{equation}
where $F_{\theta_s}(X_i)$ are the predictions of image $X_i$ via the source-pretrained model. Since this calculation is computationally intractable, we approximate it with Monte-Carlo sampling using dropout, following the method proposed by Gal and Ghahramani~\cite{DBLP:conf/icml/GalG16}. This approximation is achieved by conducting $M$ stochastic forward passes while keeping the detection model unchanged~\cite{DBLP:journals/corr/BlundellCKW15}.

Since the outputs \( F_{\theta_s}(X_i) = (\mathbf{b_i}, \mathbf{c_i}) \) consist of bounding box coordinates and class scores, the detection variance is defined as the product of the bounding box variance \( v_{bi} \) and the class score variance \( v_{ci} \). For a given prediction with \( N_i \) bounding boxes and \( K \) classes, where \( \mathbf{\{b_{ij}} = (x_{ij}^1, y_{ij}^1, x_{ij}^2, y_{ij}^2)\}_{j=1}^{N_i} \) and \( \mathbf{\{c_{ij}} = (c_{ij}^1, c_{ij}^2, \ldots, c_{ij}^K)\}_{j=1}^{N_i} \), we can express \( v_{bi} \) and \( v_{ci} \) as follows:

\begin{equation}
    v_{bi} = \frac{1}{M N_i} \sum_{j=1}^{N_i} \sum_{m=1}^{M} \| \mathbf{b_{ij}^m} - \mathbf{\bar{b}_{ij}} \|^2,
\end{equation}
\begin{equation}
    v_{ci} = \frac{1}{M N_i} \sum_{j=1}^{N_i} \sum_{m=1}^{M} \| \mathbf{c_{ij}^m} - \mathbf{\bar{c}_{ij}} \|^2,
\end{equation}
where \( \mathbf{b_{ij}^m} \) and \( \mathbf{c_{ij}^m} \) represent the localization coordinates and classification scores of the \( m \)-th forward pass for the \( j \)-th bounding box in \( X_i \), respectively, and \( \mathbf{\bar{b}_{ij}} \), \( \mathbf{\bar{c}_{ij}} \) denote their corresponding average values over all \( M \) forward passes.

The detection variance for an image \( X_i \) is computed as \( v_i = v_{bi} v_{ci} \). We then rank the images based on their variances, where \( r_i \) represents the rank of \( X_i \). The variance level \( vl_i \) for the \( i \)-th image is given by \( vl_i = \frac{r_i}{N} \). If \( vl_i \geq \sigma \), we categorize \( X_i \) as source-similar; otherwise, it is deemed source-dissimilar, with \( \sigma \in (0, 1) \) being a predefined threshold. This process effectively partitions the target domain into source-similar and source-dissimilar subsets for query-based adversarial alignment between domains.

\subsection{Query-based Domain Adversarial Learning}
\label{subsec:query}

We introduce query-based domain adversarial learning to align both easy and hard features globally. Specifically, on the encoder side, a query embedding \( q_d^{\text{enc}} \) is concatenated with the token sequence to form the input \( z_0 \) to the transformer encoder, $i.e.$,
\begin{equation}
\begin{aligned}
    {z}_{0}= \left[{q}_{{d}}^{{enc}} ; {f}_{e}^{1} ; {f}_{e}^{2} ; \cdots ; {f}_{e}^{{N}}\right]+{E}_{{pos }}+{E}_{{level}},
\end{aligned}
\end{equation}
where ${E}_{{pos }} \in \mathbb{R}^{(N+1) \times C}$ is the positional embedding, ${E}_{{level }} \in \mathbb{R}^{(N+1) \times C}$ is the feature level embedding~\cite{zhu2020deformable}.

During the encoding phase, the query adaptively aggregates domain-specific features across the entire sequence. It captures the global context from the input images and focuses more on tokens with larger domain discrepancies. This query is then passed through a domain discriminator \( D_{\text{enc}_q} \) for efficient feature representations, $i.e.$,

\begin{equation}
\mathcal{L}_{{enc}_{q}}^{\ell}=d \log {D}_{{enc}_{{q}}}\left({z}_{\ell}^{0}\right)+(1-d) \log \left(1-{D}_{{enc }_{{q}}}\left({z}_{\ell}^{0}\right)\right), 
\end{equation}
where \( \ell = 1 \ldots L_{\text{enc}} \) denotes the layer indices in the encoder, and \( d \) is the domain label, taking the value 0 for source images and 1 for target images. Similarly, we append a query \( q_{d}^{\text{dec}} \) to the object queries to form the input sequence for the transformer decoder:
\begin{equation}
    {q}_{0}=\left[{q}_{{d}}^{{dec}} ; {q}^{1} ; {q}^{2} ; \cdots ; {q}^{M}\right]+{E}_{{pos}}^{'}, \quad 
\end{equation}
where ${E}_{{pos}}^{'} \in \mathbb{R}^{\left({M}+1\right) \times C}$ is the positional embedding and ${q}^{i}$ is the $i$-th object query in the sequence.
During the decoding phase, the query combines contextual information from each object query in the sequence, capturing the relationships among objects. This query is subsequently processed through the domain discriminator \( D_{\text{dec}_k} \) to obtain more effective feature representations.

\begin{equation}
\mathcal{L}_{dec_{q}}^{\ell}= d \log {D}_{{dec}_{{k}} }\left({q}_{\ell}^{0}\right)+(1-d) \log \left(1-{D}_{{dec}_{{q}}}\left({q}_{\ell}^{0}\right)\right)
\end{equation}

where $\ell=1 \ldots L_{dec}$ indexes the layers in the transformer decoder. 

\insightbox{The query-based domain adversarial learning facilitates better feature representations in both encoder and decoder but with distinct roles—aggregating global scene layouts in the encoder and encoding object relationships in the decoder. Leveraging attention and adversarial learning, it prioritizes aligning features with significant domain gaps while minimizing effort on well-aligned ones. }


\subsection{Token-wise Domain Adversarial Learning}
\label{subsec:token}

While query-based global adversarial learning effectively bridges the global domain gap in scene layout and inter-object relationships, it faces challenges in handling domain shifts due to local textures and styles. To address this, we introduce token-wise domain adversarial learning, which is applied to both the encoder and decoder of \focalnet.

In particular, each token embedding in the encoder sequence is passed through a domain classifier \( D_{\text{enc}_k} \) for adversarial training.

\begin{equation}
\mathcal{L}_{e n c_{k}}^{\ell}=-\frac{1}{N}\sum_{i=1}^{N}\left[d \log {D}_{{enc}_{{k}}}\left({z}_{\ell}^{i}\right)+(1-d) \log \left(1-{D}_{{enc}_{{k}}}\left({z}_{\ell}^{i}\right)\right)\right].
\end{equation}
Likewise, a domain discriminator \( D_{\text{dec}_k} \) is applied on the decoder side to align each token embedding in the decoder, i.e.,

\begin{equation}
\mathcal{L}_{d e c_{k} }^{\ell}=-\frac{1}{M}\sum_{i=1}^{M}\left[d \log D_{{dec}_{k}}\left({q}_{\ell}^{i}\right)+(1-d) \log \left(1-D_{{dec}_{q}}\left({q}_{\ell}^{i}\right)\right)\right].
\end{equation}

Query-based domain adversarial learning cannot be replaced by token-wise adversarial learning. While tokens in transformers focus more on local features due to their origin from image patches or object instances, domain queries aggregate global context without focusing on local details. This allows them to better address domain gaps related to scene layout and inter-object relationships.

\insightbox{While both the \focalnet encoder and decoder use token-wise adversarial learning, the encoder focuses on local texture and appearance, while the decoder aligns domain gaps at the object-instance level.}
\subsection{Cascaded Feature Alignment}
\label{subsec:proda}

To ensure thorough feature alignment, we implement cascaded feature alignment, progressively aligning source-similar and source-dissimilar features from shallow to deep layers. In the transformer encoder, this is expressed as:
\begin{equation}
    \mathcal{L}_{enc}=\sum_{l=1}^{L_{enc}} \left(  \mathcal{L}_{enc_{k}}^{\ell} + \lambda_{enc_{q}} \mathcal{L}_{enc_{q}}^{\ell} \right),
\end{equation}
where $\lambda_{enc_{q}}$ is a hyperparameter balancing query-based and token-based alignment losses, set to 0.1 in our experiments. Similarly, in the transformer decoder, we have:
\begin{equation}
    \mathcal{L}_{dec}=\sum_{l=1}^{L_{dec}} \left( \mathcal{L}_{dec_{k}}^{\ell} + \lambda_{dec_{q}} \mathcal{L}_{dec_{q}}^{\ell} \right),
\end{equation}
with $\lambda_{dec_{q}}$ also set to 0.1. 3-layer MLPs are used as discriminators for both encoder and decoder, improving feature alignment.

\subsection{Overall Objective}
\label{subsec:total}

To summarize, the final training objective for \textbf{TITAN} is formulated as:
\begin{equation}
\min_{G} \max_{D}  \mathcal{L}_{stu}(G) - \lambda_{enc} \mathcal{L}_{enc}(G, D) - \lambda_{dec} \mathcal{L}_{dec}(G, D) \label{eq:overall}
\end{equation}
Here, \( G \) represents the object detection model, and \( D \) refers to the domain discriminators. The hyperparameters \( \lambda_{\text{enc}} \), \( \lambda_{\text{dec}} \), and \( \lambda_{\text{cons}} \) control the relative importance of the different loss terms.


\begin{table*}[t]
\centering
\caption{: Results of adaptation from normal to foggy weather (\cf). "SF" refers to the source-free setting. “Oracle” refers to the models trained by using labels during training.}
\label{t1_c2f}
\resizebox{0.8\textwidth}{!}{%
\begin{tabular}{l|l|c|cccccccc|c}

\toprule
\textbf{Method} & \textbf{Venue} & \textbf{\sf} & \textbf{Person} & \textbf{Rider} & \textbf{Car} & \textbf{Truck} & \textbf{Bus} & \textbf{Train} & \textbf{Mcycle} & \textbf{Bicycle} & \textbf{mAP} \\
\midrule
\texttt{DA-Faster}~\cite{Chen_2018_CVPR}  & CVPR'18 & \color{red}\xmark & 29.2 & 40.4 & 43.4 & 19.7 & 38.3 & 28.5 & 23.7 & 32.7 & 32.0 \\
\texttt{EPM}~\cite{hsu2020every} & ECCV'20 & \color{red}\xmark& 44.2 & 46.6 & 58.5 & 24.8 & 45.2 & 29.1 & 28.6 & 34.6 & 39.0 \\
\texttt{SSAL}~\cite{munir2021ssal} & NIPS'21 & \color{red}\xmark& 45.1 & 47.4 & 59.4 & 24.5 & 50.0 & 25.7 & 26.0 & 38.7 & 39.6 \\
\texttt{SFA}~\cite{wang2021exploring} & MM'21 & \color{red}\xmark&46.5 & 48.6 & 62.6 & 25.1 & 46.2 & 29.4 & 28.3 & 44.0 & 41.3 \\
\texttt{UMT}~\cite{Deng_2021_CVPR} & CVPR'21 &\color{red}\xmark & 33.0 & 46.7 & 48.6 & 34.1 & 56.5 & 46.8 & 30.4 & 37.3 & 41.7 \\
\texttt{D-adapt}~\cite{jiang2022decoupled} & ICLR'21 & \color{red}\xmark & 40.8 & 47.1 & 57.5 & 33.5 & 46.9 & 41.4 & 33.6 & 43.0 & 43.0 \\

\texttt{TIA}~\cite{Zhao_2022_CVPR} & CVPR'22 &\color{red}\xmark & 34.8 & 46.3 & 49.7 & 31.1 & 52.1 & 48.6 & 37.7 & 38.1 & 42.3 \\
\texttt{PT}~\cite{he2022cross} & ICML'22 &\color{red}\xmark &40.2 & 48.8 & 63.4 & 30.7 & 51.8 & 30.6 & 35.4 & 44.5 & 42.7 \\
\texttt{MTTrans}~\cite{yu2022mttrans} & ECCV'22 &\color{red}\xmark & 47.7 & 49.9 & 65.2 & 25.8 & 45.9 & 33.8 & 32.6 & 46.5 & 43.4 \\
\texttt{SIGMA}~\cite{Li_2022_CVPR} & CVPR'22 &\color{red}\xmark  &44.0 & 43.9 & 60.3 & 31.6 & 50.4 & 51.5 & 31.7 & 40.6 & 44.2 \\
\texttt{O2net}~\cite{gong2022improving} & MM'22 & \color{red}\xmark& 48.7 & 51.5 & 63.6 & 31.1 & 47.6 & 47.8 & 38.0 & 45.9 & 46.8 \\
\texttt{AQT}~\cite{huang2022aqt} & IJCAI'22 & \color{red}\xmark&49.3 & 52.3 & 64.4 & 27.7 & 53.7 & 46.5 & 36.0 & 46.4 & 47.1 \\
\texttt{AT}~\cite{li2022cross} & CVPR'22 &\color{red}\xmark &43.7 & 54.1 & 62.3 & 31.9 & 54.4 & 49.3 & 35.2 & 47.9 & 47.4 \\
\texttt{TDD}~\cite{He_2022_CVPR} & CVPR'22 & \color{red}\xmark&50.7 & 53.7 & 68.2 & 35.1 & 53.0 & 45.1 & 38.9 & 49.1 & 49.2 \\

\texttt{MRT}~\cite{zhao2023masked} & ICCV'23 & \color{red}\xmark& 52.8 & 51.7 & 68.7 & 35.9 & 58.1 & \textbf{54.5} & 41.0 & 47.1 & 51.2 \\
\texttt{HT}~\cite{deng2023harmonious} & CVPR'23 & \color{red}\xmark & 52.1 & 55.8 & 67.5 & 32.7 & 55.9 & 49.1 & 40.1 & 50.3 & 50.4 \\
\texttt{CIGAR}~\cite{liu2023cigar} & CVPR'23 & \color{red}\xmark & 45.3 & 45.3 & 61.6  & 32.1 & 50.0 & 51.0 & 31.9 & 40.4 & 44.7 \\
\texttt{CSDA}~\cite{gao2023csda} & ICCV'23 & \color{red}\xmark & 46.6 & 46.3 & 63.1 & 28.1 & 56.3 & 53.7 & 33.1 & 39.1 & 45.8 \\

\texttt{SFOD}~\cite{li2020free} & AAAI'21 & \color{blue}\checkmark & 21.7 & 44.0 & 40.4 & 32.6 & 11.8 & 25.3 & 34.5 & 34.3 & 30.6 \\
\texttt{SFOD-Mosaic}~\cite{li2020free} & AAAI'21 & \color{blue}\checkmark & 25.5 & 44.5 & 40.7 & 33.2 & 22.2 & 28.4 & 34.1 & 39.0 & 33.5 \\
\texttt{HCL}~\cite{huang2022model} & NIPS'21 & \color{blue}\checkmark & 26.9 & 46.0 & 41.3 & 33.0 & 25.0 & 28.1 & 35.9 & 40.7 & 34.6 \\
\texttt{SOAP}~\cite{soapijis21} & IJIS'21 & \color{blue}\checkmark & 35.9 & 45.0 & 48.4 & 23.9 & 37.2 & 24.3 & 31.8 & 37.9 & 35.5 \\
\texttt{LODS}~\cite{li2022source} & CVPR'22 & \color{blue}\checkmark & 34.0 & 45.7 & 48.8 & 27.3 & 39.7 & 19.6 & 33.2 & 37.8 & 35.8 \\
\texttt{AASFOD}~\cite{chu2023adversarial} & AAAI'23 & \color{blue}\checkmark & 32.3 & 44.1 & 44.6 &28.1 & 34.3 & 29.0 & 31.8 & 38.9 & 35.4 \\
\texttt{IRG}~\cite{vs2023instance} & CVPR'23& \color{blue}\checkmark & 37.4 & 45.2 & 51.9 & 24.4 & 39.6 & 25.2 & 31.5 & 41.6 & 37.1 \\
\texttt{PETS}~\cite{liu2023periodically} & ICCV'23 & \color{blue}\checkmark & 46.1 & 52.8 & 63.4 & 21.8 & 46.7 & 25.5 & 37.4 & 48.4 & 40.3 \\
\texttt{LPLD}~\cite{yoon2024enhancing} & ECCV'24 & \color{blue}\checkmark & 39.7 & 49.1& 56.6& 29.6 & 46.3 & 26.4 & 36.1 & 43.6& 40.9 \\

\midrule
\rowcolor[gray]{0.79}
\textbf{TITAN (Ours)} &  & \color{blue}\checkmark & \textbf{52.8} & \textbf{51.7} & \textbf{68.0} & \textbf{43.2} & \textbf{65.5} & 41.8 & \textbf{46.0} & \textbf{48.7} & \textbf{52.2}   \\
\midrule

Oracle &  & \color{red}\xmark &66.3 & 61.1 & 80.8 & 45.6 & 68.8 & 52.0 & 49.1 & 54.9 & 59.8  \\
\bottomrule
\end{tabular}%
}
\end{table*}

\subsection{Generalization Analysis}

Adversarial learning is applied to establish the transformation from the target domain to the source domain, and the success of this transformation is contingent on its generalization capabilities~\cite{rostamizadeh2012foundations, he2020recent}.

Let $\mu$ denote the distribution of the original data, while $\nu$ represents the distribution of the generated data. The empirical approximations of these distributions are indicated as $\hat{\mu}_N$ and $\nu_N$, where $N$ is the size of the training sample. In adversarial training, the goal is to train a generator \( g \in \mathcal{G} \) and a discriminator \( f \in \mathcal{F} \), where \( \mathcal{G} \) and \( \mathcal{F} \) are the respective hypothesis spaces. The discriminator is implemented as a multilayer perceptron (MLP) with three layers, which consists of fully connected units and non-linear activation functions, as described in Section~\ref{subsec:proda}.

\textbf{Theorem:} \textit{Covering Bound for the Discriminator}

Assume that the spectral norm of each weight matrix is bounded, i.e., \( \| A_i \|_{\sigma} \leq s_i \), and that each weight matrix \( A_i \) has a corresponding reference matrix \( M_i \) such that \( \| A_i - M_i \|_{\sigma} \leq b_i \) for \( i = 1, \dots, 3 \). Then, the covering number \( \mathcal{N} \left( \mathcal{F}|_S, \varepsilon, \|\cdot\|_2 \right) \) satisfies the following inequality:
\[
\log \mathcal{N} \left( \mathcal{F}|_S, \varepsilon, \|\cdot\|_2 \right) \leq \frac{\log \left( 2W^2 \right) \| X \|_2^2}{\varepsilon^2} \prod_{i=1}^{3} s_i^2 \sum_{i=1}^{3} \frac{b_i^2}{s_i^2},
\tag{13}
\]
where \( W \) is the maximum dimension of the feature maps in the algorithm.

This result is derived from \cite{bartlett2017spectrally, he2020resnet}, with a detailed proof provided in the Supplementary (\S~\ref{proof}). The generalizability of GANs \cite{goodfellow2014generative} depends on the complexity of the discriminator hypothesis \cite{zhang2017discrimination}. Building on this, we utilize simple discriminators to enhance both generalizability and domain adaptation performance.




%

\section{Experiments}
\label{experiments}


\subsection{Experimental Setup}

\myfirstpara{Datasets.}
The datasets used in our experiments include:
\begin{enumerate*}[label=\textbf{(\arabic*)}]
\item {Cityscapes}~\cite{cordts2016cityscapes}: This dataset comprises urban scenes with 2,975 training images and 500 validation images.
\item {Foggy Cityscapes}~\cite{sakaridis2018semantic}: Similar to Cityscapes, this dataset integrates fog and depth information into street view images.
\item {KITTI}~\cite{geiger2013vision}: A benchmark dataset for autonomous driving, containing images from real-world street scenes. For the experiments, only 7,481 training images were used.
\item {SIM10k}~\cite{johnson2016driving}: A synthetic dataset with 10,000 city scenery images of cars.
\item {BDD100k}~\cite{yu2020bdd100k}: An open-source video dataset for autonomous driving, including 100k images from various times, weather conditions, and driving scenarios. Its daytime subset consists of 36,728 training images and 5,258 validation images.
\item {RSNA-BSD1K}~\cite{carrrsna}: The original RSNA dataset~\cite{} includes 54,706 mammograms with 1,000 malignancies from 8,000 patients. RSNA-BSD1K is a subset of 1,000 images with 200 malignant cases, annotated by expert radiologists.
\item {INBreast}~\cite{moreira2012inbreast}: A smaller \bcd dataset with 410 mammography images from 115 patients, including 87 malignancies.
\item {DDSM}~\cite{lee2017curated}: A publicly available \bcd dataset comprising 2,620 full mammography images with 1162 malignancies. For full detail, please refer to the Supplementary material (\S~\ref{S_datsets}).
\end{enumerate*}

\mypara{Task Settings}
Building upon existing research~\cite{huang2021model, li2021free, zhao2023masked, oza2023unsupervised, chu2023adversarial, liu2023periodically}, we validate our method across four popular \sfdaod and \uda benchmarks. In addition to this, we perform experiments on two cross-domain Medical Imaging (\mi) tasks. These datasets represent various types of domain shifts, including\footnote{The {A}-to-{B} (\ab) notation signifies the adaptation of a model pre-trained on the source domain {A} to the target domain {B}.}:
\begin{enumerate*}[label=\textbf{(\arabic*)}]
\item {Cityscapes}-to-{Foggy-Cityscapes} (\cf)
\item {Cityscapes}-to-{BDD100k} (\cb)
\item {KITTI}-to-{Cityscapes (Car)} (\kc)
\item {Sim10k}-to-{Cityscapes (Car)} (\sc)
\item {RSNA}-to-{INBreast} (\texttt{R2In})
\item {DDSM}-to-{INBreast} (\texttt{D2In})
\end{enumerate*}
\begin{table*}[t]
\centering
\caption{Results of adaptation from small-scale to large-scale dataset (\cb).}
\label{t2_c2b}
\resizebox{0.8\textwidth}{!}{%
\begin{tabular}{l|l|c|ccccccc|c}

\toprule

\textbf{Method} & \textbf{Venue} &\textbf{\sf}& \textbf{Person} & \textbf{Rider} & \textbf{Car} & \textbf{Truck} & \textbf{Bus} & \textbf{Mcycle} & \textbf{Bicycle} & \textbf{mAP} \\
\midrule
\texttt{DA-Faster}~\cite{Chen_2018_CVPR}  & CVPR'18 & \color{red}\xmark & 28.9 & 27.4 & 44.2 & 19.1 & 18.0 & 14.2 & 22.4 & 24.9 \\
\texttt{ICR} ~\cite{xu2020exploring} & CVPR'20 & \color{red}\xmark& 32.8 & 29.3 & 45.8 & 22.7 & 20.6 & 14.9 & 25.5 & 27.4 \\
\texttt{EPM} ~\cite{hsu2020every} & ECCV'20 & \color{red}\xmark &39.6 & 26.8 & 55.8 & 18.8 & 19.1 & 14.5 & 20.1 & 27.8 \\

\texttt{SFA} ~\cite{wang2021exploring} & MM'21 & \color{red}\xmark  &40.2 & 27.6 & 57.5 & 19.1 & 23.4 & 15.4 & 19.2 & 28.9 \\
\texttt{AQT} ~\cite{huang2022aqt} & IJCAI'22 &\color{red}\xmark  &38.2 & 33.0 & 58.4 & 17.3 & 18.4 & 16.9 & 23.5 & 29.4 \\
\texttt{ILLUME} ~\cite{khindkar2022miss} & WACV'22&\color{red}\xmark & 33.2 & 20.5 & 47.8 & 20.8 & 33.8 & 24.4 & 26.7 & 29.6 \\
\texttt{O2net} ~\cite{gong2022improving} & MM'22 & \color{red}\xmark& 40.4 & 31.2 & 58.6 & 20.4 & 25.0 & 14.9 & 22.7 & 30.5 \\
\texttt{AWADA}~\cite{menke2022awada} & arXiv'22 &\color{red}\xmark &41.5 & 34.2 & 56.0 & 18.7 & 20.0 & 20.4 & 29.7 & 31.5 \\
\texttt{MTTrans} ~\cite{yu2022mttrans} & ECCV'22 & \color{red}\xmark& 44.1 & 30.1 & 61.5 & 25.1 & 26.9 & 17.7 & 23.0 & 32.6 \\
\texttt{MRT} ~\cite{zhao2023masked} & ICCV'23 &\color{red}\xmark& 48.4 & 30.9 & 63.7 & 24.7 & 25.5 & 20.2 & 22.6 & 33.7 \\
\texttt{SFOD}~\cite{li2020free} & AAAI'21 & \color{blue}\checkmark & 31.0 & 32.4 & 48.8 & 20.4 & 21.3 & 15.0 & 24.3 & 27.6 \\
\texttt{SFOD-M}~\cite{li2020free} & AAAI'21 & \color{blue}\checkmark & 32.4 & 32.6 & 50.4 & 20.6 & 23.4 & 18.9 & 25.0 & 29.0 \\
\texttt{PETS}~\cite{liu2023periodically} & ICCV'23 & \color{blue}\checkmark & 42.6 & 34.5 & 62.4 & 19.3 & 16.9 & 17.0 & 26.3 & 31.3 \\
\texttt{AASFOD}~\cite{chu2023adversarial} & AAAI'23 & \color{blue}\checkmark & 33.2 & 36.3 & 50.2 & 26.6 & 24.4 & 22.5 & 28.2 & 31.6 \\

\midrule
\rowcolor[gray]{0.87}
\textbf{TITAN (Ours)}  &&\color{blue}\checkmark  &\textbf{49.9} &\textbf{35.6} &\textbf{65.7}  & \textbf{24.6}  &\textbf{35.9}  &\textbf{31.5}  &29.2 &\textbf{38.3}  \\
\midrule
Oracle  & &   &69.2 &51.3 & 83.1 & 62.9 & 63.3 &49.6  &50.0 &61.3  \\
\bottomrule
\end{tabular}%
}
\end{table*}

\mypara{Implementation Details}
We adopt FocalNet-DINO (\fnd)~\cite{yang2022focal} as our base detector. Initially, we set the loss coefficients to $\lambda_\text{enc} = 1.0$ and $\lambda_{dec} = 0.9$. The weight smoothing parameter $\alpha$ in the Exponential Moving Average (\ema) is configured to be $0.9996$. The network is optimized using the Adam optimizer~\cite{kingma2014adam} with an initial learning rate of $2 \times 10^{-4}$ and a batch size of $8$. For data augmentation, we employ random horizontal flipping for basic augmentation and apply more advanced techniques such as random color jitter, grayscaling, and Gaussian blurring. The implementation is done using PyTorch~\cite{paszke2017automatic}. The min-max loss function is realized through gradient reversal layers~\cite{ganin2016domain}. Further details are provided in Supplementary (\S~\ref{S_hyper}).

\mypara{Evaluation Metrics}
For natural image datasets, we report the average precision (\ap) for each individual class along with the mean average precision (\map) score, consistent with previous studies. For medical image datasets, we utilize the Free-Response Receiver Operating Characteristic (\froc) curves to evaluate detection performance, alongside F1-score and AUC for classification results. The \froc curves visually represent the trade-off between sensitivity/recall and false positives per image (\fpi). We consider a prediction to be a true positive if the center of the predicted bounding box falls within the ground truth box ~\cite{rangarajan2023deep}.

\begin{table}[t]
\centering
\caption{\textcolor{red}{(Left)} Results of adaptation from synthetic to real scenes (\sc). \textcolor{red}{(Right)} Results of adaptation across cameras (\kc).
}
\label{t3_city}
\resizebox{\columnwidth}{!}{%
\begin{tabular}{l|l|c|c|c}
\cmidrule{1-5} 
\multicolumn{3}{c}{\texttt{setting}}
& Sim2City (\sc)
& Kitty2City (\kc)
  \\
\cmidrule{1-5} 
\textbf{Method} & \textbf{Venue} & \textbf{\texttt{SF}} & \textbf{Car (AP)}&\textbf{Car (AP)} \\
\cmidrule{1-5} 

\texttt{DA-Faster}~\cite{Chen_2018_CVPR} & CVPR'18 &  \color{red}\xmark & 41.9& 41.8 \\

\texttt{SAPNet}~\cite{li2020spatial} & ECCV'20 & \color{red}\xmark&  44.9&  43.4 \\
\texttt{EPM}~\cite{hsu2020every} & ECCV'20 & \color{red}\xmark & 49.0 & 43.2 \\
\texttt{GPA}~\cite{xu2020exploring} & CVPR'20 & \color{red}\xmark & 47.6& 47.9 \\


\texttt{MeGA-CDA}~\cite{vs2021megacda}& CVPR'21 &  \color{red}\xmark & 44.8& 43.0 \\



\texttt{DSS}~\cite{wang2021domainspecific} & CVPR'21 & \color{red}\xmark&  44.5&  42.7\\

\texttt{ViSGA}~\cite{rezaeianaran2021seeking}& ICCV'21 & \color{red}\xmark & 49.3 & 47.6\\

\texttt{SFA}~\cite{wang2021exploring}& MM'21 &  \color{red}\xmark& 52.6& 41.3\\


\texttt{PT}~\cite{he2022cross} & ICML'22 & \color{red}\xmark&  -  & 55.1\\
\texttt{MTTrans}~\cite{yu2022mttrans} & ECCV'22 & \color{red}\xmark&  57.9  & -\\
\texttt{LODS}~\cite{li2022source} & CVPR'22 & \color{red}\xmark&  -  & 43.9\\
\texttt{CIGAR}~\cite{liu2023cigar} & CVPR'23 & \color{red}\xmark& 58.5& 48.5\\
\texttt{CSDA}~\cite{gao2023csda} & ICCV'23 & \color{red}\xmark&  57.8  & 48.6\\

\texttt{SFOD}~\cite{li2020free} & AAAI'21 & \color{blue}\checkmark &  42.3 &  43.6\\
\texttt{SFOD-Mosaic}~\cite{li2020free} & AAAI'21 & \color{blue}\checkmark &  42.9 &  44.6\\
\texttt{IRG}~\cite{vs2023instance} & CVPR'21 & \color{blue}\checkmark&  45.2 &  46.9\\
\texttt{AASFOD}~\cite{chu2023adversarial} & AAAI'23 & \color{blue}\checkmark&  44.0&  44.9 \\
\texttt{PETS}~\cite{liu2023periodically} & ICCV'23 & \color{blue}\checkmark&  57.8 & 47.0 \\ 
\texttt{LPLD}~\cite{yoon2024enhancing} & ECCV'24 & \color{blue}\checkmark&  49.4 & 51.3 \\ 
\cmidrule{1-5} 

\cellcolor[gray]{0.87}\textbf{ (Ours)} &\cellcolor[gray]{0.87}  &\cellcolor[gray]{0.87} \color{blue}\checkmark&\cellcolor[gray]{0.87} \textbf{59.8}  &\cellcolor[gray]{0.87}\textbf{53.2} \\
\cmidrule{1-5} 

Oracle & &\color{red}\xmark &63.9 & 62.1 \\
\cmidrule{1-5} 
\end{tabular}%
}

\end{table}

\subsection{Comparison with Current \sota Methods}

We evaluate the performance of our proposed \textbf{TITAN} method against other approaches on the four natural benchmarks and two medical benchmarks mentioned earlier. Since \uda and \sfdaod share similar task settings, we conducted comparisons with both. Table \ref{t1_c2f}-\ref{t3_city} and Table \ref{t4_breast} present the comparison results on natural and medical images, respectively. Our proposed \textbf{TITAN} consistently outperforms existing state-of-the-art (\sota) methods, demonstrating significant improvements across both natural and medical images, highlighting its effectiveness in both domains and its potential to advance performance in real-world applications. This further establishes \textbf{TITAN} as a promising solution for tackling domain adaptation challenges in various tasks.

\begin{table}[t]
\centering

  \resizebox{\columnwidth}{!}{
    \begin{tabular}{c|l|l|c|cccc|cc}
    \toprule
    \textbf{Exp}& \textbf{Method} & \textbf{Venue} & \textbf{SF} &\textbf{R@0.05}  & \textbf{R@0.3} &\textbf{R@0.5}& \textbf{R@1.0}  & \textbf{AUC} & \textbf{F1-score} \\ 
    \midrule
    & D-adapt~\cite{jiang2022decoupled} & ICLR'21 & \color{red}\xmark & 0.04 &0.12 &0.18 &0.29 &0.439&0.263\\
    & AT~\cite{li2022crossdomain} & CVPR'22 &\color{red}\xmark& 0.16 &0.28 &0.35 &0.42 &0.486&0.338\\
    & H2FA~\cite{9878659} & CVPR'22 &\color{red}\xmark& 0.03 &0.13 &0.18 &0.36 &0.634&0.236\\
    \texttt{R2In} & MRT~\cite{zhao2023masked} & ICCV'23 & \color{red}\xmark & 0.32 &0.52 &0.69 &0.72 &0.741&0.352\\
    & Mexformer~\cite{wang2021exploring} & MM'21 &\color{blue}\checkmark& 0.24 &0.31 &0.39&0.39 &0.336&0.287\\
    & IRG~\cite{deng2021unbiased} & CVPR'23&\color{blue}\checkmark&0.16 &0.25 &0.37&0.39&0.308& 0.235\\
    & LPLD~\cite{yoon2024enhancing} & ECCV'24&\color{blue}\checkmark&0.25 &0.25 &0.45&0.43&0.548& 0.635\\
    \midrule
    \rowcolor[gray]{0.87}
    & Ours  &  &&\textbf{0.59} &\textbf{0.78} &\textbf{0.80}&\textbf{0.83}&\textbf{0.892}& \textbf{0.850}\\
    \midrule
    \midrule
    \textbf{Exp}& \textbf{Method} & \textbf{Venue} & \textbf{SF} &\textbf{R@0.05}  & \textbf{R@0.3} &\textbf{R@0.5}& \textbf{R@1.0}  & \textbf{AUC} & \textbf{F1-score} \\ 
    \midrule
    & D-adapt~\cite{jiang2022decoupled} & ICLR'21 & \color{red}\xmark & 0.00 &0.06 &0.09 &0.1 &0.381&0.362\\
    & AT~\cite{li2022crossdomain} & CVPR'22 &\color{red}\xmark& 0.01 &0.08 &0.10 &0.15 &0.385&0.311\\
    & H2FA~\cite{9878659} & CVPR'22 &\color{red}\xmark& 0.02 &0.08 &0.10 &0.12 &0.483&0.315\\
    \texttt{D2In} & MRT~\cite{zhao2023masked} & ICCV'23 &\color{red}\xmark & 0.03 &0.09 &0.12 &0.17 &0.739&0.587\\
    & Mexformer~\cite{wang2021exploring} & MM'21 &\color{blue}\checkmark& 0.02 &0.03 &0.03&0.03 &0.06&0.09\\
    & IRG~\cite{deng2021unbiased} & CVPR'23 &\color{blue}\checkmark&0.05 &0.05 &0.07&0.09&0.11& 0.12\\
    & LPLD~\cite{yoon2024enhancing} & ECCV'24&\color{blue}\checkmark&0.09 &0.15 &0.35&0.35&0.548& 0.635\\
    \midrule
    \rowcolor[gray]{0.87}
    & Ours  &  &&\textbf{0.36} &\textbf{0.51} &\textbf{0.75}&\textbf{0.81}&\textbf{0.825}& \textbf{0.838}\\
    \bottomrule
    \end{tabular}
      }
  \label{tab:sota_uda_comparison}

\caption{\textcolor{red}{(left-top)} Results on adaptation from large to small-scale medical datasets with different modalities (\texttt{R2In}), \textcolor{red}{(left-bottom) }Results on adaptation across medical datasets with different machine-acquisitions (\texttt{D2In})}
\label{t4_breast}
\end{table}
\subsubsection{Adaptation on Natural Images}

\begin{figure*}
    \centering
    \includegraphics[width=\linewidth]{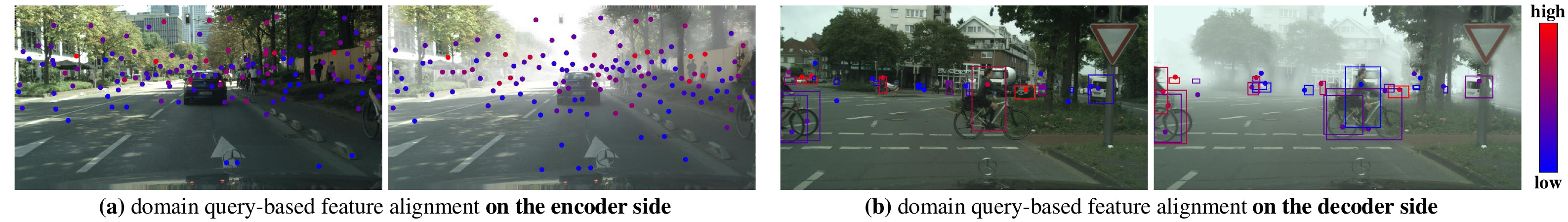}
    \caption{"Visualization of domain queries from both the encoder and decoder sides of \focalnet, in the \cf scenario."}
    \label{fig:query}
\end{figure*}

\begin{figure}
    \centering
    \includegraphics[width=\linewidth]{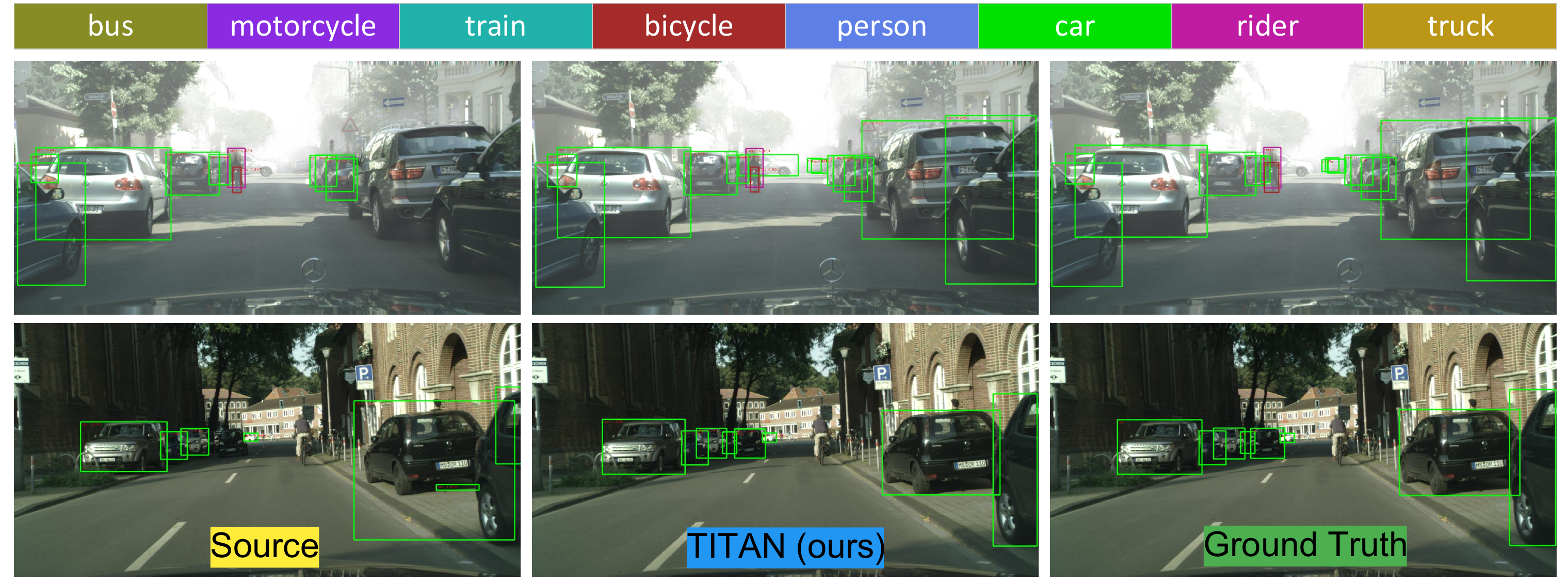}
    \caption{Qualitative results comparing Source model (\focalnet), \textbf{TITAN} (ours), and ground truth visualizations. Our method shows accurate predictions.}
    \label{fig:vis}
\end{figure}

\begin{enumerate}[leftmargin=*]
\item (\cf): Adaptation results from clear to foggy weather shown in \cref{t1_c2f}. Our method surpasses both \uda and \sfdaod methods.
\item (\cb): Adaptation from small to large-scale datasets. Please see \cref{t2_c2b}.
\item (\kc): Adaptation across different cameras. \cref{t3_city} evaluates our model's performance on domain shifts due to different camera settings viz. resolution, quality etc. 
\item (\sc): Adaptation from synthetic to real images. To explore adaptation from synthetic to real scenes, we employ a model pretrained on the complete \textbf{SIM10k} dataset~\cite{johnson2016driving} as the source model. We then adapt this model to the \textbf{Cityscapes} dataset~\cite{cordts2016cityscapes}, with only car images retained and other categories discarded as shown in \cref{t3_city}. Further details are provided in the Supplementary (\S~\ref{S_datsets}).
\end{enumerate}

\subsubsection{Adaptation on Medical Imaging Datasets}

\begin{enumerate}[leftmargin=*]
\item (\texttt{R2In}): Adapting across datasets with different machine acquisitions, using RSNA-BSD1K~\cite{carrrsna} as source and INBreast~\cite{moreira2012inbreast} as the target. Our method outperforms existing approaches across \fpi values (\cref{t4_breast}).  
\item (\texttt{D2In}): Adapting from large to small-scale medical datasets with different modalities, from \textbf{DDSM}~\cite{lee2017curated} to \textbf{INBreast}~\cite{moreira2012inbreast}. Our method achieves \sota results across \fpi values (\cref{t4_breast}).

\end{enumerate}

\subsection{Ablation Studies}
\label{ssec:ablation}
To gain deeper insights into our method, we perform ablation studies by isolating each component of \textbf{TITAN}, as presented in Table~\ref{tab:more_ablation}. We have the following observations: (1) both domain query-based adversarial learning and token-wise adversarial learning can alleviate the domain gaps and improve \focalnet transformer's cross-domain performance by 40.8 and 39.7 mAP, respectively; (2) query-based domain adversarial learning and token-wise domain adversarial learning are complementary to each other. Thereby, a combination of both brings further improvement; (3) The target division (TD) based adversarial learning is effective, resulting in a gain of 51.2 mAP.
\begin{table}[t]
\renewcommand\arraystretch{0.9}
\centering
\small
\caption{Ablation studies on query-token based domain adversarial learning, \textbf{without target division}, are conducted on the \cf scenario. $\text{DQ}_{enc}$ and $\text{DQ}_{dec}$ are applied to the final encoder and decoder layers, respectively. Similarly, $\text{TW}_{enc}$ and $\text{TW}_{dec}$ denote \textbf{TITAN} applied to the last encoder and decoder layers.} 
\label{tab:more_ablation}
\setlength{\tabcolsep}{2mm}
\resizebox{\columnwidth}{!}{%
    \begin{tabular}{cccccc|c}
        \toprule
        Baseline & $\text{DQ}_{enc}$ & $\text{DQ}_{dec}$ & $\text{TW}_{enc}$ & $\text{TW}_{dec}$& TD&  mAP \\
        \midrule
        $\color{blue}\checkmark$ &  &  &  &   & &35.2 \\
       \midrule
        \multirow{8}{*}{} & $\color{blue}\checkmark$ &  &&  &  &  37.5 \\
        &  & $\color{blue}\checkmark$ &  &  & & 36.7 \\
        &  &  & $\color{blue}\checkmark$ &  & & 38.5 \\
        &  &  &  & $\color{blue}\checkmark$ & &37.9 \\
        & $\color{blue}\checkmark$ & $\color{blue}\checkmark$ &  &  && 40.8 \\
        &  &  & $\color{blue}\checkmark$ & $\color{blue}\checkmark$& &  39.7 \\
        & $\color{blue}\checkmark$ &  & $\color{blue}\checkmark$ & & & 47.1 \\
        &  & $\color{blue}\checkmark$ &  & $\color{blue}\checkmark$ &&  45.6 \\
        \midrule
        $\color{blue}\checkmark$& $\color{blue}\checkmark$ & $\color{blue}\checkmark$ & $\color{blue}\checkmark$ & $\color{blue}\checkmark$ & \color{blue}\checkmark &  \textbf{51.2} \\
        \bottomrule
    \end{tabular}%
}
\vspace{3mm}
\end{table}


\subsubsection{Visualization of Domain Query.}

The query-token domain feature alignment adaptively aggregates global context in the encoder and decoder. As visualized in Fig.~\ref{fig:query}, the encoder query attends to regions with domain gaps, like dense fog, while the decoder query emphasizes foreground objects such as cars and bicycles, capturing key domain shifts. 
More visualizations and ablations are provided in the Supplementary Material (\S~\ref{qualitative}).

\subsubsection{Detection Results}
Fig.~\ref{fig:vis} presents in-depth visual comparisons between \focalnet and our framework \textbf{TITAN}, alongside the ground-truth annotations. As shown, \textbf{TITAN} consistently enhances detection performance across all three scenarios. It effectively reduces false positives and identifies difficult objects that \focalnet misses. In the first row, \textbf{TITAN} successfully detects a car that is distant and not labeled in the ground-truth, highlighting its strong generalization capability to the target domain.

\section{Conclusion}
\label{conclusion}

We have proposed a novel source-free domain adaptive object detection framework that divides the target domain into easy and hard samples, and later aligns them in the feature space using the query-token domain advesarial learning.  Through comprehensive experiments conducted across various natural and medical image benchmark datasets, we consistently demonstrate that our technique outperforms state-of-the-art. Our ablation studies confirm the effectiveness of each design aspect in improving the model's domain adaptation capability.

{
    \small
    \bibliographystyle{ieeenat_fullname}
    \bibliography{main}
}

\clearpage
\setcounter{page}{1}
\maketitlesupplementary

\renewcommand{\thesection}{\AlphAlph{\value{section}-5}} 

\textit{\textbf{Sec.~\ref{proof}} provides additional theoretical insights.}\\

\textit{\textbf{Sec.~\ref{S_datsets}} provides additional details on training datasets.}\\

\textit{\textbf{Sec.~\ref{S_hyper}} provides additional details on hyperparamers.}\\

\textit{\textbf{Sec.~\ref{S_hyper}} provides further architectural details.}\\

\textit{\textbf{Sec.~\ref{s_backbone}} provides details about multiple backbones.}\\

\textit{\textbf{Sec.~\ref{S_augmentation}} provides additional details on augmentations.}\\

\textit{\textbf{Sec.~\ref{qualitative}} provides extensive  ablation and visualizations.}

\section{Detailed Theoretical Analysis}
\label{proof}
\begin{proof}[Proof of Preposition]

We denote the spaces of the output functions $F_{(A_1, \dots, A_{i-1})}$ induced by the weight matrices $A_i, i = 1, \dots, 5$ by $\mathcal{H}_i, i = 1, \dots, 5$, respectively. Lemma A.7 in \cite{bartlett2017spectrally} suggests the following inequality:
\begin{equation}
\resizebox{\columnwidth}{!}{$
\begin{aligned}
    \log \mathcal{N}(\mathcal{F}|S) 
    &\le \log \left( \prod_{i=1}^{5} \sup_{\mathbf{A}_{i-1} \in \bm{\mathcal{B}}_{i-1}} \mathcal{N}_i \right) \\
    &\le \sum_{i=1}^{5} \log \left(
        \sup_{\substack{(A_1, \dots, A_{i-1}) \\ \forall j < i, A_j \in B_j}} 
        \mathcal{N} \left( \left\{ A_i F_{(A_1, \dots, A_{i-1})} \right\},
        \varepsilon_i, \| \cdot \|_2 \right)
    \right)
\end{aligned}
$}
\tag{1}
\end{equation}

Thus, we obtain the following inequality:
\begin{align}
    \log \mathcal{N}(\mathcal{F}|S) 
    &\le \sum_{i=1}^{5} \frac{b_i^2 \| F_{(A_1, \dots, A_{i-1})}(X) \|_2^2}{\varepsilon_i^2} \log \left( 2W^2 \right) ~.
     \tag{2}
\end{align}

Meanwhile, we have the following inequality:
\[
\resizebox{\columnwidth}{!}{$
\begin{aligned}
    \| F_{(A_1, \dots, A_{i-1})}(X) \|_2^2 
    &= \| \sigma_{i-1} (A_{i-1} F_{(A_1, \dots, A_{i-2})}(X)) - \sigma_{i-1}(0) \|_2 \\
    &\le \| \sigma_{i-1} \| \| A_{i-1} F_{(A_1, \dots, A_{i-2})}(X) - 0 \|_2 \\
    &\le \rho_{i-1} \| A_{i-1} \|_{\sigma} \| F_{(A_1, \dots, A_{i-2})}(X) \|_2 \\
    &\le \rho_{i-1} s_{i-1} \| F_{(A_1, \dots, A_{i-2})}(X) \|_2.
\end{aligned}
$}
\tag{3}
\]

Therefore, we get:
\begin{equation}
    \| F_{(A_1, \dots, A_{i-1})}(X) \|_2^2 
    \le \|X\|^2 \prod_{j=1}^{i-1} s_i^2 \rho_i^2.
     \tag{5}
\end{equation}

Motivated by the proof in \cite{bartlett2017spectrally}, we assume the following equations:
\begin{align}
    \varepsilon_{i+1} &= \rho_i s_{i+1} \varepsilon_i \nonumber \\
    \varepsilon_5 &= \rho_1 \prod_{i=2}^4 s_i \rho_i s_5 \epsilon_1 \nonumber \\
    \varepsilon &= \rho_1 \prod_{i=2}^5 s_i \rho_i \epsilon_1 
    \tag{6}
\end{align}

Therefore, we have:
\begin{equation}
    \varepsilon_i = \frac{\rho_i \prod_{j=1}^{i-1} s_j \rho_j}{\prod_{j=1}^5 s_j \rho_j} \varepsilon ~.
     \tag{7}
\end{equation}

Thus, we obtain:
\begin{align}
\log \mathcal{N}(\mathcal{F}|_S, \varepsilon, \| \cdot \|_2) 
\le \frac{\log\left( 2W^2 \right) \| X \|_2^2}{\varepsilon^2} \left( \prod_{i=1}^5 s_i \rho_i \right)^2 \sum_{i=1}^5 \frac{b_i^2}{s_i^2}
\tag{8}
\end{align}
which is precisely Equation (13).
The proof is completed.
\end{proof}

\section{Additional Details on Training datasets}
\label{S_datsets}
Our experiments are conducted on five \sfdaod benchmark datasets. Along with this, we introduce the \sfdaod problem in the medical domain and conduct experiments on four Breast Cancer Detection (\bcd) datasets, including two publicly available datasets. 
\begin{table*}[htbp]
\caption{Details of Training Datasets}
\label{tab:training_datasets}
\centering
\resizebox{\textwidth}{!}{%
\begin{tabular}{c|c|c|c|p{2.2 cm}|c}
\toprule
\textbf{Dataset} & \textbf{Type} & \multicolumn{2}{c|}{\textbf{Pre-training (Source)}} & \multicolumn{2}{c}{\textbf{Unsupervised Adaptation (Target)}} \\
\midrule
& & \textbf{Train} & \textbf{Val/Test} & \centering\textbf{Train}  & \textbf{Test} \\ \midrule
Cityscapes & Natural & 2,975 & 500 & \centering 2,975  & 500 \\ 
Foggy Cityscapes & Natural & - & - & \centering 2,975 & 500 \\ 
KITTI & Natural & 7,481 &  & \centering - & - \\
SIM10k & Natural & 8,500 &1,500 & \centering - & - \\ 
BDD100k & Natural &  - & - & \centering 36,728 & 5,258 \\ \midrule
RSNA-BSD1K & Medical & 1000 (200) & 250 (50) &\centering 1000 (200) & 1000(200) \\ 
INBreast & Medical & - & - &\centering 410 (91) & 410 (91) \\ 
DDSM & Medical & 2885 (1339) & 218 (118) & \centering3103 (1458) &3103 (1458) \\ 

\bottomrule
\end{tabular}} 
\end{table*}

\subsection{Natural Datasets}
1) \textbf{Cityscapes}~\cite{cordts2016cityscapes} is gathered from urban environments across 50 European cities, provides detailed annotations for 30 semantic classes across 8 categories. It comprises 5,000 high-quality annotated images and a larger set of 20,000 coarsely annotated images, all high-resolution (2048x1024 pixels) providing detailed visual data for precise scene understanding. Our study utilizes the high-quality subset of 5,000 images, consisting of 2,975 training images and a standard test set of 500 images from Frankfurt, Munster, and Lindau, as employed in prior research. This dataset is release under the Creative Commons Attribution-NonCommercial-ShareAlike 4.0 International License, and is available for academic and research purposes.

2) \textbf{Foggy Cityscapes}~\cite{sakaridis2018semantic} is an extension of the Cityscapes dataset, designed to support research in developing robust computer vision algorithms for autonomous driving in foggy conditions. Similarly, it consists of high-resolution images (2048x1024 pixels) with annotations ofr 2D bounding boxes, pixel-level semantic segmentation, and instance segmentation, inherited from the original dataset. It is directly constructed from Cityscapes by simulating three levels of foggy weather (0.005,0.001,0.02), but we adapt on the most extreme level (0.02) in our experiments as done by \cite{sakaridis2018semantic}. This dataset is released under the Creative Commons Attribution-NonCommercial-ShareAlike 4.0 International License.

3) \textbf{KITTI}~\cite{geiger2013vision} is a promenient benchmark for computer vision and robotics, particularly in autonomous driving. Data was collected using a car-mounted sensor suite, including high-resolution color and grayscale cameras, a Velodyne laser scanner, and a GPS/IMU system, in urban, rural, and highway settings around Karlsruhe, Germany. The dataset includes stereo image pairs with disparity maps, consecutive frames for optical flow, image sequences with ground truth poses for SLAM, and images with corresponding 3D point clouds for object detection. The images are typically high-resolution (1242x375 pixels), captured at 10-100 Hz, providing detailed and diverse visual data. The 3D object detection subset is split into 7,481 training images and 7,518 test images. As quite common in the Domain Adaptation literature, we only use the 7,481 training images in our experiments for adaptation and evaluate on the same split as done in . The dataset is publicly available for academic use under the Creative Commons Attribution-NonCommercial-ShareAlike license.  

4) \textbf{SIM10k}~\cite{johnson2016driving} is a synthetic dataset create using the Grand Theft Auto V (GTA V) engine, it simulates various driving scenarios providing diverse set of high-resolution images with detailed annotations. This dataset consists of 10,0000 images of urban environments under different weather conditions, lighting and traffic situations. In our experiments for source training, we prepare our own train and val set of 8,500 and 1,500 images respectively. We intend to make this split public for reproducibility. This dataset is also publicly available for academic and research purposes, with usage terms typically provided by the dataset creators, allowing for non-commercial use.

5) \textbf{BDD100k}~\cite{yu2020bdd100k} developed by the Berkeley DeepDrive (BDD) team, is one of the largest and most diverse driving video datasets available. It comprises of 100,000 driving videos and 100K keyframe images, and captures a wide range of driving scenarios across urban, suburban and rural environments in the United States, under diverse weather conditions, lighting, and times of day. Each image is of 720p and is richly annotated with 2D bounding boxes for objects, lane markings, drivable areas and scene attributes like weather and time of day. We make use of the standard BDD100K train set containing 36,728 images for adaptation and use 5,258 images for evaluation. These images are the frames at the 10th second in the videos and the split is consistent with the original video set. This dataset is released under the Creative Commons Attribution-Non Commercial-ShareAlike 4.0 Internation license and is publicly available for academic and research purposes.
\subsection{Medical Datasets}

\textbf{INBreast}~\cite{moreira2012inbreast} is a relatively small breast cancer detection dataset, consisting of 410 mammography images from 115 patients, including 87 confirmed malignancies. The training images include both histologically confirmed cancers and benign lesions initially recalled for further examination but later identified as nonmalignant. We hypothesize that incorporating both malignant and benign lesions in the training process will enhance our model's ability to detect a broader range of lesions and effectively distinguish between malignant and benign cases.

\textbf{DDSM}~\cite{lee2017curated} is a publicly available breast cancer detection dataset, comprising 2,620 full mammography images, with 1,162 containing malignancies. The DDSM dataset offers digitized film-screen mammography exams with lesion annotations at the pixel level, where cancerous lesions are histologically confirmed. We used the DDSM dataset exclusively for training our model and not for evaluation. This decision stems from the observation that the quality of digitized film-screen mammograms is inferior to that of full-field digital mammograms, making evaluation on these cases less relevant. For our purposes, we converted the lossless JPEG images to PNG format, mapped pixel values to optical density using calibration functions from the DDSM website, and rescaled the pixel values to a range of 0–255.

\textbf{RSNA-BSD1K}~\cite{carrrsna}  is a comprehensive collection of 54,706 screening mammograms sourced from approximately 8,000 patients. This dataset includes a diverse range of cases, among which 1,000 instances have been identified as malignant. The dataset serves as a valuable resource for developing and evaluating machine learning models in the field of medical imaging, particularly in breast cancer detection. From this large dataset, a specialized subset is curated known as RSNA-BSD1K, which consists of 1,000 carefully selected mammograms. This subset was designed to maintain a balance between normal and malignant cases while ensuring high-quality annotations suitable for robust model training and evaluation. Within RSNA-BSD1K, 200 cases have been confirmed as malignant, representing a diverse spectrum of tumor characteristics and imaging conditions.

Note that unlike in natural images, single domain detection techniques which use a particular subset of the
dataset for adaptation and remaining for testing, our technique does not require
any labels from the target dataset. Hence, for medical datasets, it seems logical to
use the whole dataset during training and testing, and not just the any
train or test split. Hence, when reporting results for “Dataset A to Dataset B”,
we imply that the model is trained on $\mathcal{D}_s = A$ (whole dataset for the training),
and adapted for $\mathcal{D}_t = B$ (whole dataset for adaptation in an unsupervised way and testing). Table X shows the detailed split wise sets used during experiments.

\section{Further Insights into Hyperparameter Selection}
The selection of hyperparameters, as detailed in Table~\ref{tab:implementation_details1}, was guided by empirical experimentation and domain-specific considerations. Key factors included optimizing model generalization, ensuring stability during training, and balancing performance across different benchmarks. Parameters such as the number of pseudo labels, learning rate, and loss weights were fine-tuned based on validation results, with adjustments made dynamically for specific dataset shifts. Additionally, threshold values for pseudo-labeling were set adaptively to enhance robustness across diverse datasets.
\label{S_hyper}
\begin{table}[H]
    \centering
    \caption{Below are the detailed hyper-parameters corresponding to each benchmark, with the source dataset as the In-house dataset}
    \vspace{0.1em}
    \label{tab:implementation_details1}
    \resizebox{0.5\textwidth}{!}{  
    \begin{tabular}{@{}clc@{}}
        \toprule
        \textbf{Hyper-parameter} & \textbf{Description} & \textbf{Value}  \\
        \midrule
        \textit{num\_classes} & Number of classes & 1 \\
        \textit{lr} & Learning rate & 0.0001 \\
        \textit{lr\_backbone} & Learning rate for backbone & 1e-05 \\
        \textit{batch\_size} & Batch size & 4 \\
        \textit{weight\_decay} & Weight decay & 0.0001 \\
        \textit{epochs} & Number of epochs & 100 \\
        \textit{lr\_drop} & Learning rate drop & 11 \\
        \textit{clip\_max\_norm} & Clip max norm & 0.1 \\
        \textit{multi\_step\_lr} & Multi-step learning rate & True \\
        \textit{modelname} & Model name & 'dino' \\
        \textit{backbone} & Backbone & 'focalnet\_L\_384\_22k\_fl4' \\
        \textit{focal\_levels} & Focal levels & 4 \\
        \textit{focal\_windows} & Focal windows & 3 \\
        \textit{position\_embedding} & Position embedding & 'sine' \\
        \textit{pe\_temperature} & PE temperature & 20 \\
        \textit{enc\_layers} & Encoder layers & 6 \\
        \textit{dec\_layers} & Decoder layers & 6 \\
        \textit{dim\_feedforward} & Dimension of feedforward network & 2048 \\
        \textit{hidden\_dim} & Hidden dimension & 256 \\
        \textit{dropout} & Dropout & 0.0 \\
        \textit{nheads} & Number of heads & 8 \\
        \textit{num\_queries} & Number of queries & 900 \\
        \textit{box\_attn\_type} & Box attention type & 'roi\_align' \\
        \textit{num\_feature\_levels} & Number of feature levels & 4 \\
        \textit{enc\_n\_points} & Encoder points & 4 \\
        \textit{dec\_n\_points} & Decoder points & 4 \\
        \textit{transformer\_activation} & Transformer activation & 'relu' \\
        \textit{batch\_norm\_type} & Batch norm type & 'FrozenBatchNorm2d' \\
        \textit{set\_cost\_class} & Set cost class & 2.0 \\
        \textit{set\_cost\_bbox} & Set cost bbox & 5.0 \\
        \textit{set\_cost\_giou} & Set cost GIoU & 2.0 \\
        \textit{cls\_loss\_coef} & Class loss coefficient & 1.0 \\
        \textit{mask\_loss\_coef} & Mask loss coefficient & 1.0 \\
        \textit{dice\_loss\_coef} & Dice loss coefficient & 1.0 \\
        \textit{bbox\_loss\_coef} & BBox loss coefficient & 5.0 \\
        \textit{giou\_loss\_coef} & GIoU loss coefficient & 2.0 \\
        \textit{enc\_loss\_coef} & Encoder loss coefficient & 1.0 \\
        \textit{focal\_alpha} & Focal alpha & 0.25 \\
        \textit{matcher\_type} & Matcher type & 'HungarianMatcher' \\
        \textit{nms\_iou\_threshold} & NMS IoU threshold & 0.1 \\
        \textit{use\_dn} & Use DN & True \\
        \textit{dn\_number} & DN number & 100 \\
        \textit{dn\_box\_noise\_scale} & DN box noise scale & 1.0 \\
        \textit{dn\_label\_noise\_ratio} & DN label noise ratio & 0.5 \\
        \textit{dn\_labelbook\_size} & DN labelbook size & 3 \\
        \textit{use\_ema} & Use EMA & False \\
        \textit{ema\_decay} & EMA decay & 0.9997 \\
        \textit{optim\_iter\_per\_epoch} & Optimization iterations per epoch & 2500 \\
        \bottomrule
    \end{tabular}}
\end{table}

\begin{table*}[t]
    \centering
    \caption{Detailed hyper-parameters corresponding to each benchmark, with the source dataset as the In-house dataset.}
    \vspace{0.1em}
    \label{tab:implementation_details}
    \begin{adjustbox}{width=\textwidth}
    \begin{tabular}{@{}clcccccc@{}}
        \toprule
        \textbf{Hyper-parameter} & \textbf{Description} & \textbf{City2Foggy}  & \textbf{City2BDD} & \textbf{Sim2City} & \textbf{Kitti2City} & \textbf{InH2InB} & \textbf{InH2DDSM} \\
        \midrule
        \textit{num\_classes} & Number of classes & 8 & 2 & 2 & 2 & 2 & 2 \\
        \textit{epochs} & Number of epochs & 10 & 10 & 10 & 10 & 20 & 20  \\
        \textit{topk\_pseudo} & Number of pseudo labels & 30 & 30 & 30 & 30 & 15 & 15 \\
        \textit{use\_dynamic\_th} & Use dynamic threshold & True & True & True & True & False & False \\
        \textit{pseudo\_th} & Initial pseudo label threshold & 0.006 & 0.04 & 0.04 & 0.04 & 0.06 & 0.06 \\
        \textit{lambda} & Weight of Enc and Dec loss & 0.6 & 0.6 & 0.6 & 0.6 & 1.0 & 1.0 \\
        \bottomrule
    \end{tabular}
    \end{adjustbox}
\end{table*}

\section{More details on different Backbones}
\label{s_backbone}
\textbf{RN50\_4scale and RN50\_5scale}. ResNet-50 is a 50-layer deep convolutional neural network designed for image recognition tasks. It employs residual learning to allow the network to learn residual functions with reference to the layer inputs, which helps in training deeper networks. The model is pretrained on the ImageNet-1k dataset, which contains 1.2 million images and 1,000 classes. This pretraining helps the network to learn robust feature representations that can be fine-tuned for various downstream tasks. Since some methods adopt 5 scales of feature maps
and some adopt 4, we report our results with both 4 and 5 scales of feature maps.

\textbf{Convnext}
This is a modern convolutional network architecture that aims to bridge the gap between convolutional networks and vision transformers. ConvNeXt incorporates design principles from transformers, such as a simplified architecture with fewer layers and parameters, but retains the efficiency and scalability of convolutional networks. It leverages state-of-the-art techniques like LayerScale, deep supervision, and various normalization methods to achieve competitive performance on benchmarks. ConvNeXt models are often pretrained on large datasets like ImageNet-1k or ImageNet-22k to provide strong initial weights for transfer learning.

\textbf{Swin}
The Swin Transformer, particularly the large version (SwinL), is a hierarchical vision transformer designed for image classification, object detection, and segmentation tasks. It introduces the concept of shifted windows for computing self-attention, which allows it to handle varying scales of features more efficiently than traditional transformers. SwinL is pretrained on the ImageNet-22k dataset, which contains 14 million images and 22,000 classes. This extensive pretraining helps the model to capture rich and diverse feature representations, making it highly effective for a wide range of visual tasks.
\begin{table}[t]
\centering
\caption{Data Augmentation Methods and Parameters}
\label{tab:augmentation}
\resizebox{\columnwidth}{!}{%
\begin{tabular}{c|c|c}
\toprule
\textbf{Augmentation Type} & \textbf{Method} & \textbf{Parameters} \\ \midrule
\multirow{3}{*}{Weak} 
  & Random Horizontal Flip & Probability = 0.5 \\ \cmidrule{2-3}
  & Resize & Size = 800, Max Size = 1333 \\ \midrule
\multirow{8}{*}{Strong} 
  & \multirow{4}{*}{Color Jitter (Color Adjustment)} 
  & Brightness = 0.4 \\ \cmidrule{3-3}
  & & Contrast = 0.4 \\ \cmidrule{3-3}
  & & Saturation = 0.4 \\ \cmidrule{3-3}
  & & Hue = 0.1 \\ \cmidrule{2-3}
  & Random Grayscale & Probability = 0.2 \\ \cmidrule{2-3}
  & \multirow{2}{*}{Gaussian Blur} 
  & Sigma Range = [0.1, 2.0 \\ \cmidrule{3-3}
  & & Probability = 0.5\\ \cmidrule{2-3}
  & \multirow{2}{*}{Normalization} 
  & Mean = [0.485, 0.456, 0.406] \\ \cmidrule{3-3}
  & & Std = [0.229, 0.224, 0.225] \\ \bottomrule
\end{tabular}%
}
\end{table}

\textbf{FN-Fl3  and FN-Fl4}
At the core of Focal Modulation Networks (FocalNets) is the focal modulation mechanism: A lightweight element-wise multiplication as the focusing operator to allow the model to see or interact with the input using the proposed modulator;  As depicted below, the modulator is computed with a focal aggregation procedure in two steps: focal contextualization to extract contexts from local to global ranges at different levels of granularity and gated aggregation to condense all context features at different granularity levels into the modulator. We adopt the same stage configurations and hidden dimensions as those used in Focal Transformers, but we replace the Self-Attention (SA) modules with Focal Modulation modules. This allows us to construct various Focal Modulation Network (FocalNet) variants. In FocalNets, we only need to define the number of focal levels (L) and the kernel size (k) at each level. For simplicity, we increase the kernel size by 2 for each subsequent focal level, i.e., \( k_{l} = k_{l-1} + 2 \). To match the complexities of Focal Transformers, we design both small receptive field (SRF) and large receptive field (LRF) versions for each of the four layouts by using 3 and 4 focal levels, respectively. We use non-overlapping convolution layers for patch embedding at the beginning (with a kernel size of 4 × 4 and stride of 4) and between stages (with a kernel size of 2 × 2 and stride of 2).

The results of utilizing each backbone in our model and the corresponding results are present in Table X.

\section{More Details on Augmentations}
\label{S_augmentation}

In our study, we employed both weak and strong augmentation techniques to enhance the robustness and generalization capabilities of our model. These augmentations are applied to the training images to simulate various real-world scenarios and improve the model's performance.

\subsection*{Weak Augmentation}

\begin{figure*}[t]
    \centering
    \includegraphics[width =0.9\textwidth, height = 17 cm]{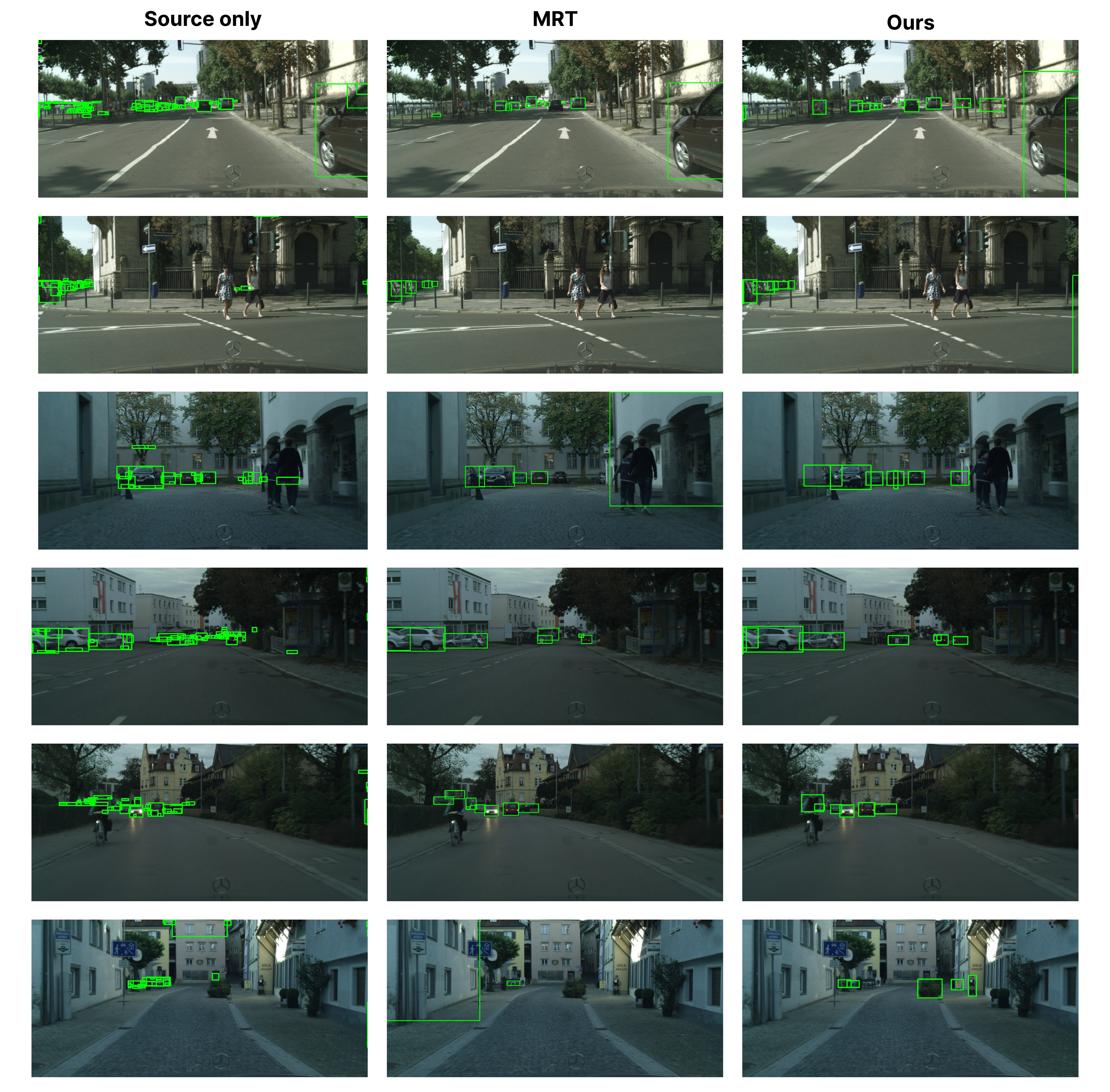}
    \caption{Qualitative results for car detection on Cityscapes in \sc setting. \texttt{MRT} is an Unsupervised domain adaptation technique with the best results in \sc. Our method, being an \sfdaod technique, comes surprisingly close to the best performing \texttt{UDA} method.}
    \label{fig:enter-label}
\end{figure*}
Weak augmentations are relatively simple transformations that slightly alter the images without significantly changing their content. We utilized two primary weak augmentation techniques. The first is \textit{Random Horizontal Flip}, which flips the image horizontally with a probability of 0.5. This helps the model learn invariance to the left-right orientation of objects, making it more robust to such variations. The second technique is \textit{Resize}, where images are resized such that the shortest edge is 800 pixels while maintaining the aspect ratio, and if the longest edge exceeds 1333 pixels, the image is scaled down accordingly. This resizing standardizes the input image size, ensuring consistent and efficient training. Together, these weak augmentations provide a baseline level of variability in the training data, helping the model to generalize better across different image scales and orientations.

\subsection*{Strong Augmentation}

Strong augmentations involve more complex transformations that significantly alter the images, thereby providing a broader range of variability. These augmentations are designed to challenge the model and improve its ability to handle diverse and complex real-world scenarios. The first strong augmentation is \textit{Color Jitter}, which randomly changes the brightness, contrast, saturation, and hue of the images with specified parameters (brightness = 0.4, contrast = 0.4, saturation = 0.4, hue = 0.1) and a probability of 0.8. This helps the model become invariant to different lighting conditions and color variations. The second augmentation is \textit{Random Grayscale}, which converts the image to grayscale with a probability of 0.2, encouraging the model to focus on shapes and structures rather than colors. The third strong augmentation is \textit{Gaussian Blur}, applied with a sigma range of [0.1, 2.0] and a probability of 0.5, simulating out-of-focus conditions and reducing high-frequency noise. Finally, all images are converted to tensors and normalized using the mean [0.485, 0.456, 0.406] and standard deviation [0.229, 0.224, 0.225]. These strong augmentations introduce substantial variability in the training data, forcing the model to learn more robust and generalized features.

\section{Qualitative Analysis}
\label{qualitative}
\begin{figure*}[t]
    \centering
    \includegraphics[width =\textwidth, height = 18cm]{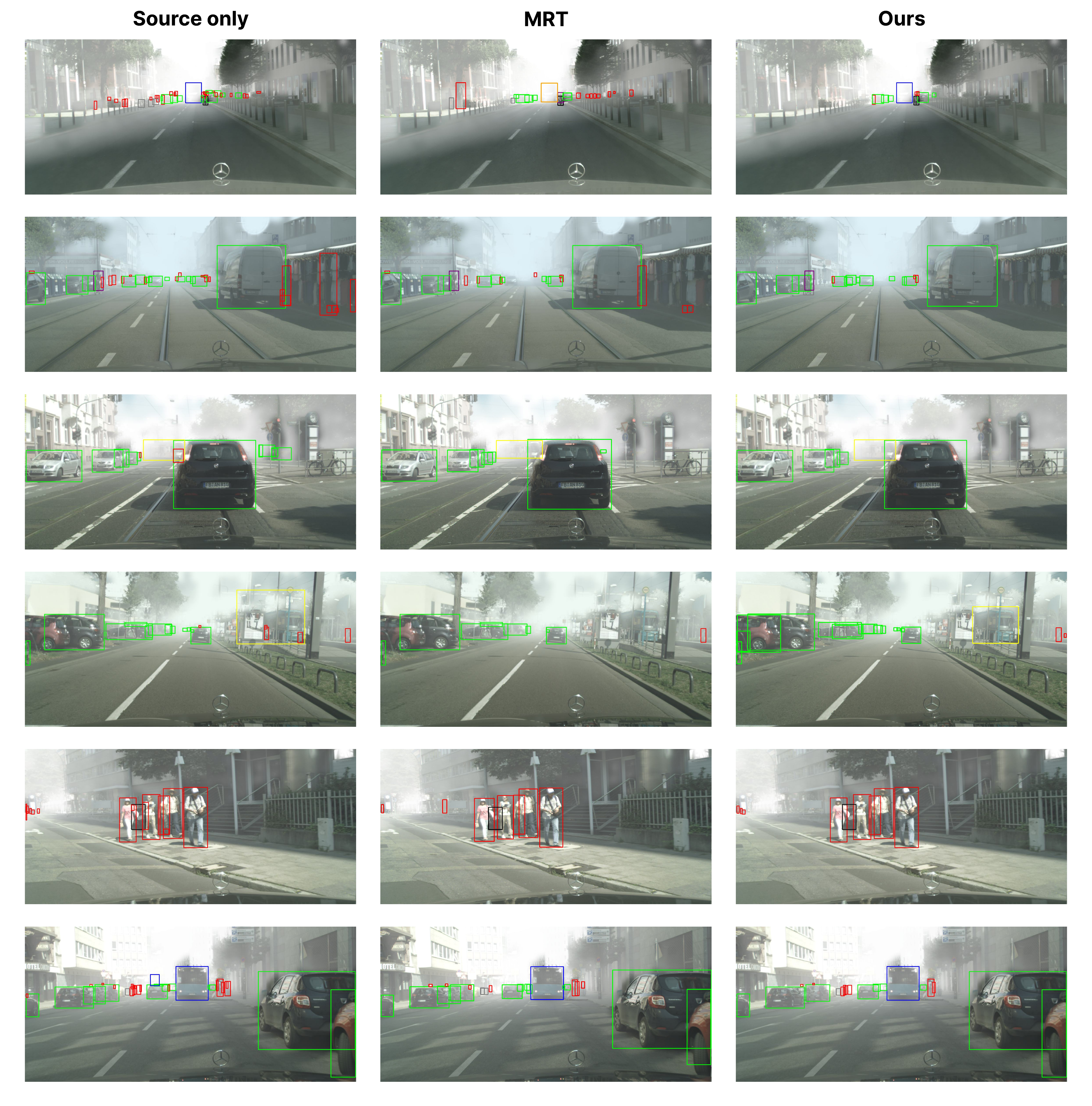}
    \caption{Qualitative results for car detection on Foggy Cityscapes in \cf setting. The visible detections are from the classes: \textcolor{red}{Person}, \textcolor{green}{Car}, \textcolor{yellow}{Train}, \textcolor{gray}{Bicycle}, \textcolor{blue}{Bus}, \textcolor{orange}{Truck}, \textcolor{black}{motorcycle}, \textcolor{violet}{Rider}. \texttt{MRT} is an Unsupervised domain adaptation technique with the best results in \cf. Our method, being an \sfdaod technique, comes surprisingly close to the best performing \texttt{UDA} method.}
    \label{fig:foggy_vis}
\end{figure*}

\begin{figure*}[t]
    \centering
    \includegraphics[width =\textwidth, height= 5 cm]{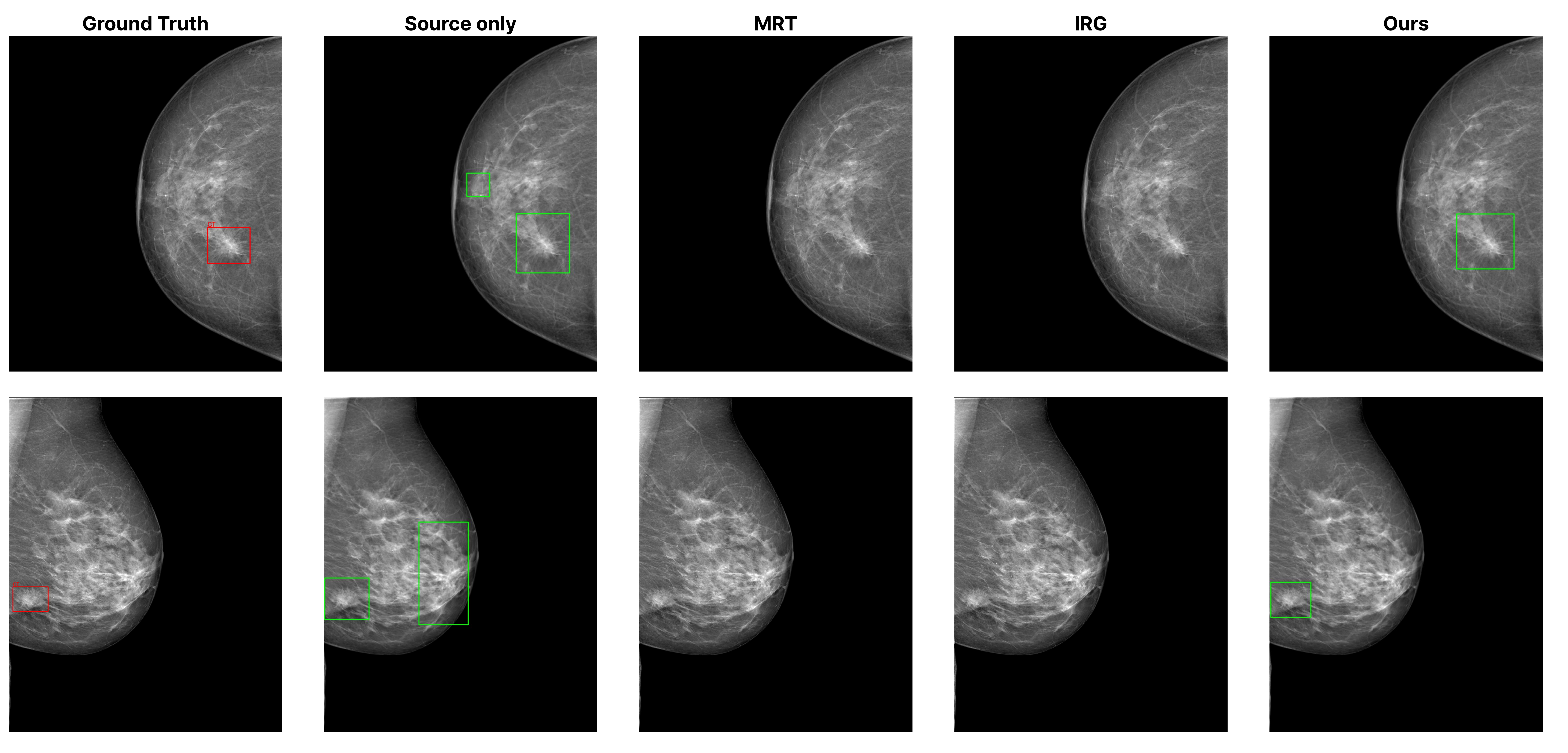}
    \caption{Quantitative results for Breast Cancer detection in \ii setting. Predictions are depicted in \textcolor{green}{green}.}
    \label{fig:vis_breast}
\end{figure*}

We provide a qualitative visualization analysis of pseudo-labels and feature distributions. Our method outperforms the state-of-the-art methods, including MRT~\cite{zhao2023masked} and GT. Figure \ref{fig:vis_breast} and Fig.~\ref{fig:foggy_vis} present predictions on the breast cancer dataset and foggy dataset respectively, highlighting the superior performance of our approach. 
\

\end{document}